\documentclass[runningheads]{llncs}
\usepackage{graphicx}

\usepackage{tikz}
\usepackage{comment}
\usepackage{amsmath,amssymb}
\usepackage{color}

\usepackage[accsupp]{axessibility}

\usepackage{times}
\usepackage{soul}
\usepackage{url}
\usepackage[hidelinks]{hyperref}
\usepackage[utf8]{inputenc}
\usepackage[small]{caption}
\usepackage{graphicx}

\usepackage{booktabs}
\usepackage{algorithm}
\usepackage{algorithmic}
\usepackage{epsfig}
\usepackage{subfiles}
\usepackage{color,soul}
\usepackage{caption}
\usepackage{subcaption}
\usepackage{array}
\usepackage[english]{babel}
\usepackage{comment}
\usepackage{multirow}
\usepackage{adjustbox}
\usepackage{boldline}
\newcolumntype{P}[1]{>{\centering\arraybackslash}p{#1}}
\newcolumntype{M}[1]{>{\centering\arraybackslash}m{#1}}
\usepackage{arydshln}

\usepackage{abdalmageed_style}

\begin{document}

\pagestyle{headings}
\mainmatter
\def\ECCVSubNumber{23}  

\title{SW-VAE: Weakly Supervised Learn Disentangled Representation Via Latent Factor Swapping}

\titlerunning{SW-VAE}

\author{Jiageng Zhu\inst{1,2,3} \and Hanchen Xie\inst{2,3} \and Wael Abd-Almageed\inst{1,2,3}}
\authorrunning{J. Zhu, H. Xie and W. Abd-Almageed}

\institute{USC Ming Hsieh Department of Electrical and Computer Engineering \and
USC Information Sciences Institute \and Visual Intelligence and Multimedia Analytics Laboratory\\
\email{\{jiagengz, hanchenx, wamageed\}@isi.edu}}

\maketitle

\newcommand{\algmse}{\textbf{SW-VAE}$_{SIM}$}
\newcommand{\alggan}{\textbf{SW-VAE}$_{GAN}$}
\newcommand{\alg}{\textbf{SW-VAE}}

\definecolor{betaVAE_color}{RGB}{170, 69, 69}
\definecolor{AnnealedVAE_color}{RGB}{170, 121, 69}
\definecolor{FactorVAE_color}{RGB}{111, 170, 69}
\definecolor{DIPVAE_color}{RGB}{69, 170, 138}
\definecolor{TCVAE_color}{RGB}{69, 138, 170}
\definecolor{AdaVAE_color}{RGB}{69, 96, 170}
\definecolor{SWVAE_color}{RGB}{20, 5, 105}
\newcommand{\eg}{\emph{e.g.,}} 
\def\Eg{\emph{E.g.,}}
\def\ie{\emph{i.e.,}} \def\Ie{\emph{I.e}}
\def\cf{\emph{c.f.,}} \def\Cf{\emph{C.f}}
\def\etc{\emph{etc}} \def\vs{\emph{vs}}
\def\wrt{w.r.t} \def\dof{d.o.f}
\def\etal{\emph{et al.}}

\begin{abstract}
Representation disentanglement is an important goal of the representation learning that benefits various of downstream tasks. To achieve this goal, many unsupervised learning representation disentanglement approaches have been developed. However, the training process without utilizing any supervision signal have been proved to be inadequate for disentanglement representation learning. Therefore, we propose a novel weakly-supervised training approach, named as \textbf{SW-VAE}, which incorporates pairs of input observations as supervision signal by using the generative factors of datasets.
Furthermore, we introduce strategies to gradually increase the learning difficulty during training to smooth the training process. As shown on several datasets, our model shows significant improvement over state-of-the-art (SOTA) methods on representation disentanglement tasks.
\end{abstract}

\section{Introduction}
\label{sec:intro}

Deep neural network (DNN) has achieved great success in many computer vision tasks, such as image classification \cite{5206848}, face recognition \cite{6909616} and image generation \cite{goodfellow2014generative}.
Learning a latent representation $\boldsymbol{z}$ from input data $\boldsymbol{x}$ is a critical first step in training DNNs, in which the objective is to learn lower dimensional representations that facilitate downstream tasks, including classification \cite{he2015deep}, few-shot learning \cite{NIPS2016_90e13578} or semantic segmentation \cite{7298965}.

Two of the fundamental challenges for robust latent representations are \emph{overfitting} and \emph{interpretability.} 
Since DNNs are massively paramterized models, they require large amounts of training data that sufficiently span all \emph{factors of variations}, such as pose, expression and gender in face recognition models \cite{liu2015faceattributes}. In the absence of large scale dataset that spans all factors of variations, DNNs tend to overfit to the underlying factors of variations. Moreover, since the training objective is minimizing the empirical risk, standard methods for training DNNs (\eg~ \cite{simonyan2015deep,kingma2014autoencoding}) produce latent representations that lack semantic meaning and hard to interpret without further processing.

\begin{figure*}
\centering
\includegraphics[width=0.6\textwidth]{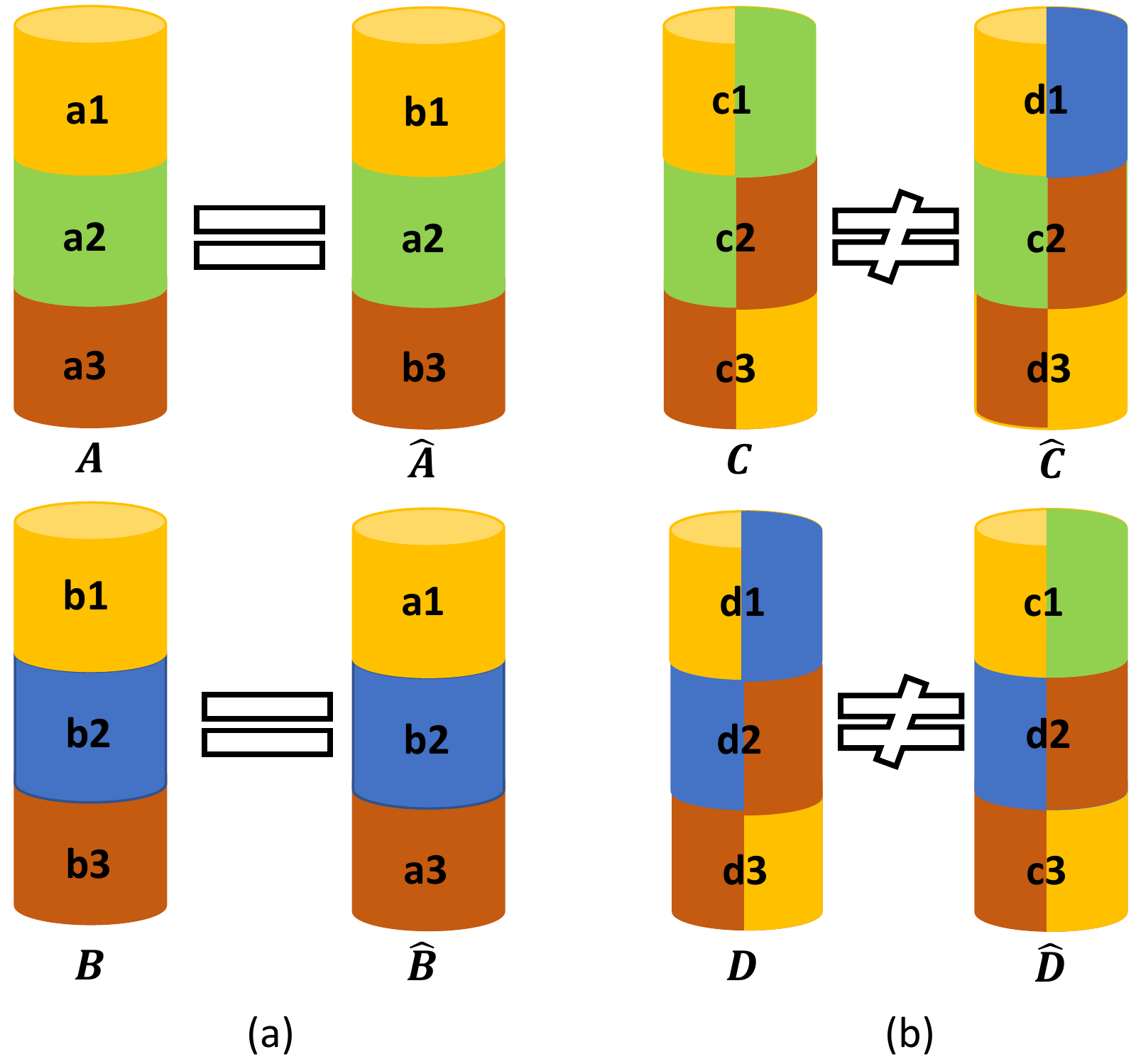}
 \caption{The basic concept of swapping latent factors of variation, represented by different colors. Left (a): If two representations are completely disentangled (\eg~$A$ and $B$), then swapping equivalent latent factors (a1 with b1, and a3 with b3) will lead to similar representations. Right (b): If $C$ and $D$ are entangled, then swapping latent factors (c1 with d1 and c3 with d3) may lead to dissimilar representations.}
     \label{fig:fig1}
\end{figure*}

Bengio \etal~\cite{6472238} define a disentangled representation $\boldsymbol{z}$ such that a change in a given representation dimension $z_i$ corresponds to a change in \emph{one and only one} underlying factor of variation $i$ of the data (\ie~invariant to all other factors). The definition of \cite{6472238} addresses the semantic interpretability of the representation in standard training methods, and is used as the basis for methods such as Factor Variational Autoencoder (FactorVAE) \cite{pmlr-v80-kim18b} and $\beta$-VAE \cite{DBLP:conf/iclr/HigginsMPBGBML17}. Recent work revealed the benefits of disentangled representations of factors of variations in various downstream tasks such as, visual reasoning~\cite{sjoerd19}, interpretability~\cite{6472238,DBLP:conf/iclr/HigginsMPBGBML17}, filtering out nuisance~\cite{kumar2018}, answering counterfactual questions~\cite{siddharth2017learning}, and fairness~\cite{DBLP:journals/corr/abs-1906-02589}.

Unsupervised disentanglement methods \cite{pmlr-v80-kim18b,DBLP:conf/iclr/HigginsMPBGBML17,chen2019isolating,Hinton504}
 relying on variational autoencoder (VAE) \cite{kingma2014autoencoding} which assume that the latent representation follows a normal distribution, where the encoder is used to estimate the posterior $p_{\theta}(\mathbf{z}|\mathbf{x})$ and the decoder is used to estimate $p_{\phi}(\mathbf{x}|\mathbf{z})$. The loss function is constructed for comparing the difference between prior and posterior. However, Locatello \etal~\cite{locatello2019challenging} proved that there are an infinite number of entangled models whose latent representation $z$ has the same marginal distribution with the ideal disentangled model, and since unsupervised learning methods only use information of the observations $\mathbf{x}$, they can not discriminate between the disentangled model and other entangled models. Locatello \etal~\cite{locatello2019challenging} further empirically analyzed state-of-the-art (SOTA) unsupervised models on different datasets, which demonstrates the necessity of supervision signal. To address the challenge, many semi-supervised and weakly supervised learning methods have been proposed \cite{pmlr-v119-locatello20a,Locatello2020Disentangling,DBLP:conf/aaai/ChenB20}. One of these is methods is Ada-VAE \cite{pmlr-v119-locatello20a} which requires the knowledge of exact number of different generative factors between a pair of images in order to achieve disentangled representation with guarantees. Rather than relying on such precise prior knowledge, our method only requires the information of the maximum number of different generative factors, and empirical results demonstrate that the disentangled representation can still be achieved with weaker supervision signal.

In this paper, we introduces a new method for learning latent representations with disentangled factors of variations via introducing weak supervision signals during training to encourage the model to learn disentangled representations. Inspired by the self-supervised learning methods which utilize the supervision provided by pairs of inputs \cite{gidaris2018unsupervised,chen2020simple}, we propose using pairs of inputs to introduce supervision signals. After encoding input pairs into latent representations, similar to the swapping method used in image manipulation \cite{park2020swapping}, we encourage the disentanglement by swapping the latent factors and comparing their corresponding reconstructions with the original inputs. As the simple example illustrated in \Cref{fig:fig1}, when the representation is fully disentangled, swapping the same elements does not change their distribution, whereas the distribution of entangled representation changes after swapping. 
In disentanglement learning, DSD \cite{NEURIPS2018_fdf1bc56}, which also adopts swapping concept, is trained under two steps. Firstly, a pair of labeled inputs are encoded, swapped, and decoded. In the second step, a random $k$-th part of latent representations encoded from unlabelled inputs is swapped and decoded. Therefore, DSD is trained under semi-supervised condition, and as discussed in the DSD \cite{NEURIPS2018_fdf1bc56}, more than $20\%$ of labeled data are needed to train the model. However, in many cases, the number of labeled data is limited due to the annotation cost is expensive. Compared to DSD, our method does not require actual label information while merely needing access to the total number of generative factors, and we empirically prove that our method achieves comparable performance. On the occasion of the exact number of different generative factors in the pairs are available, where the supervision strength is still weaker than DSD, our method can outperform DSD. Furthermore, we propose training strategies which progressively adjust the difficulty of task during learning process.

Our contributions in this paper are: (1) \alg, a new weakly supervised representation disentanglement framework and (2) extensive quantitative and qualitative experimental evaluation demonstrating that ~\alg~ outperforms SOTA representation disentanglement methods on various datasets including dSprites \cite{dsprites17}, 3dshapes \cite{3dshapes18}, MPI3D-toy \cite{gondal2019transfer}, MPI3D-realistic \cite{gondal2019transfer}, and MPI3D-real \cite{gondal2019transfer}. 
\section{Related Work}
\label{sec:related_work}

\textbf{Learning Disentanglement Representation:} Variational autoencoders (VAE) \cite{kingma2014autoencoding} is the basic framework of most SOTA disentanglement methods. VAE uses a DNN to map the inputs into latent representation modeled as a distribution denoted by $q_{\phi}(z|x)$. Recent methods modifies VAE by adding implicit or explicit regularization to disentanglement representation.  $\beta$-\textbf{VAE} \cite{DBLP:conf/iclr/HigginsMPBGBML17} adds and tunes a hyper-parameter $\beta$ before KL divergence  ($D_{KL}$) in order to achieve balanced performance on both reconstruction and latent representation disentanglement. 
In $\beta$-VAE, when $\beta>1$, the distribution $q_{\phi}(z|x)$ is used to calculate disentanglement regularization by comparing with the assuming prior $p(z)$. Kullback-Leibler divergence ($D_{KL}$) is used to calculate the distance between $q_{\phi}(z|x)$ and $p(z)$ for disentanglement regularization. The disentanglement regularization is then combined with the quality of image reconstruction to be the total objective function of the model training. 

\begin{figure*}
\centering
\includegraphics[width=\linewidth]{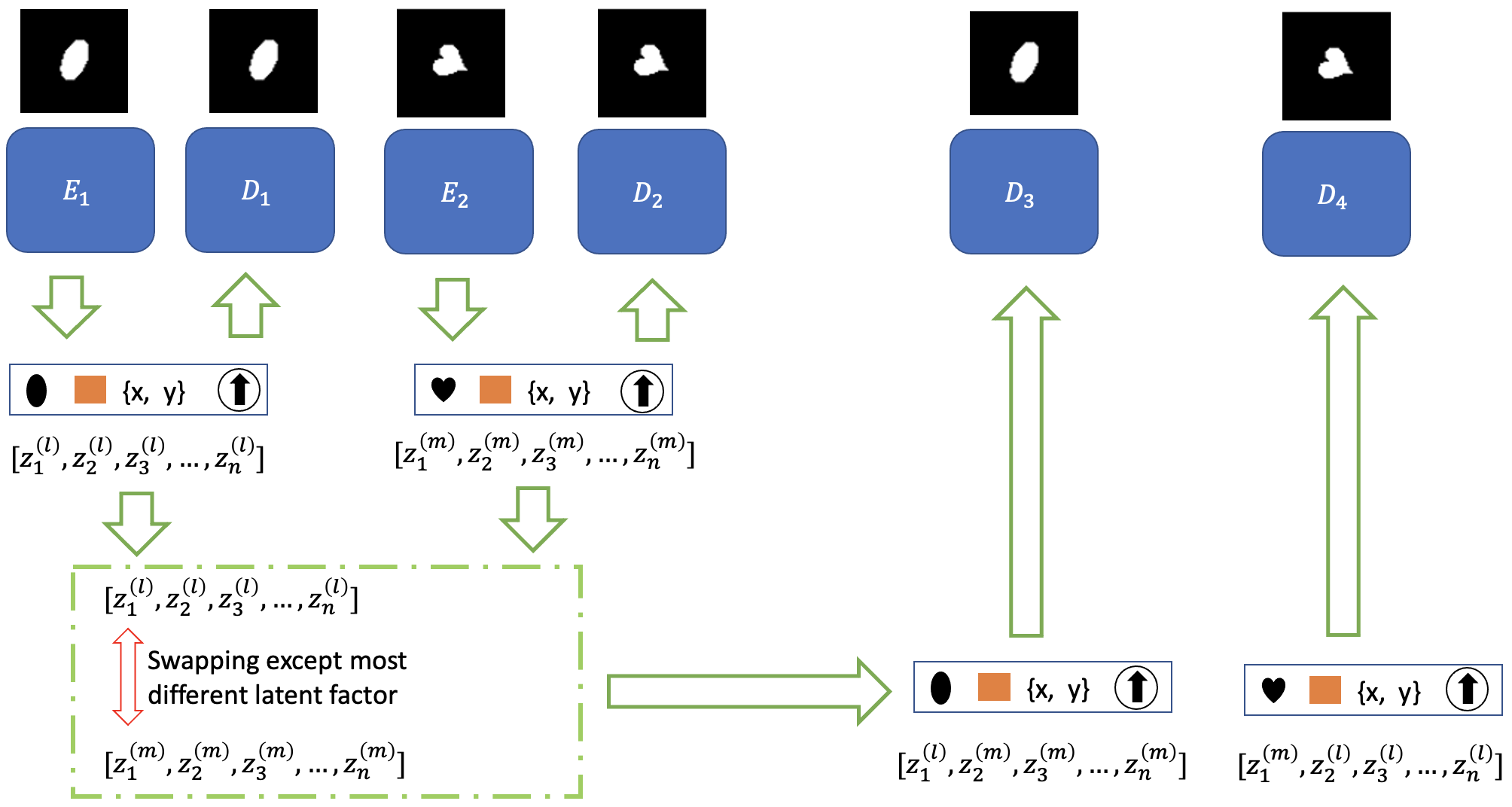}
\caption{Framework of \textbf{SW-VAE}, $E_1$ and $E_2$ share same parameters; $D_1$,$D_2$,$D_3$,$D_4$ share same parameters}
\label{fig:algorithm1_2}
\end{figure*}

Burgess \etal~\cite{burgess2018understanding} proposed \textbf{AnnealedVAE}, a modification of $\beta$-\textbf{VAE}. Intuitively, \textbf{AnnealedVAE} can be viewed as gradually increasing the capacity of latent encoding. When the encoding capacity is low, the model is forced only to encode input data which brings the most significant improvement in reconstructions. As capacity increases during training, the model learns to encode other semantic factors into the latent representation progressively while continues to disentangle the previous learned factors. 
\textbf{FactorVAE} \cite{pmlr-v80-kim18b} proposes a better trade-off between reconstruction quality and disentanglement by incorporating the discriminator to calculate the Total Correlation (TC) between $q(z)$ and $\prod q(z_i)$. The discriminator in \textbf{FactorVAE} has the same function as the discriminator in \textbf{GAN} \cite{goodfellow2014generative}.  \textbf{DIP-VAE} \cite{kumar2018} adds $D(q_{\phi}(z)||p(z))$ as additional regularization to encourage disentanglement, where $q_{\phi}(z)$ is the marginal distribution of the latent representation $z$ learned by model and $q_{\phi}(z)=\int q_{\phi}(z|x)p(x) dx$. $D$ stands for any suitable distance function.
$\beta$-\textbf{TCVAE} \cite{chen2019isolating} decomposes the $D_{KL}$ regularization used in $\beta$-\textbf{VAE} into three parts: index-code mutual information, total correlation and dimension-wise KL divergence. Index-code mutual information controls the mutual information between input data and factors in latent representation. Total correlation encourages the model to find the statistically independent factors in latent space and dimension-wise individual latent components are kept from overly diverging from their priors via KL divergence.

\textbf{Weakly supervised learning:}
Zhou \etal~\cite{10.1093/nsr/nwx106} conclude that there are three forms of weak supervision in general. The first type is incomplete supervision, \ie~only part of training samples have labels. This condition can be addressed by semi-supervised learning or active learning. The second type is inexact supervision where the labels are less precise than labels used in supervised learning, or only coarse-grained labels are provided. The third type is inaccurate supervision. The labels of some data are wrong in this case, \eg~some images or languages are labelled into the wrong class.

\section{SW-VAE Model}
\label{sec:methods}

As discussed in \Cref{sec:intro}, Bengio \etal~\cite{6472238} define representation disentanglement as learning a latent representation $z \in \mathbb{R}^{d_z}$ of input  observation $x\in \mathbb{R}^{d_x}$, in which each dimension $z_i$ changes \emph{if and only if} an underlying factor of variation $i$ of the data changes, and therefore the joint probability can be modeled as $p(z)=\prod_{i=1}p(\mathbf{z_i})$. Meanwhile, each latent element $z_i$ is expected to contain one and only one semantic meaning, such that traversing alone one latent element and fixing other elements will only change one factor of variantion within the reconstructed images obtained from the decoder \cite{kingma2014autoencoding}. Thus, for each factor of variation, there is only one highly associated latent factor. When measuring the performance of different disentanglement models, this principle is used to evaluate both the degree of disentanglement and the level of similarity between a latent factor and one semantic generative factor.

The number of generative factors of variations is less than the dimension of latent representation ($v\le d_z$), such that a subset of $z$ encodes information that is irrelevant to the generative factors of variations, and does not necessarily have semantic meaning; yet still satisfying the independency assumptions.  Disentanglement therefore leads to a latent factor $z_i$ that is \emph{invariant}  to all other factors of variation of $x$.

In variational autoencoder (VAE) \cite{DBLP:journals/corr/KingmaW13}, a variational model $q_{\phi}(z|x)$ is used to produce a probability distribution $q_\phi$ given an input sample $x$. This essentially simulates sampling a latent representation $z$ from a prior distribution $p_{\theta}(z)$, where $\theta$ and $\phi$ are the generative and variational parameter spaces respectively. The overall loss function of the VAE is shown in \Cref{eq:bvae}. 
\begin{equation} 
\label{eq:bvae}
L_{VAE}(x,z)  = -\mathbb{E}_{q_{\phi}(z|x)}[logp_{\theta}(x|z)]  + \beta D_{KL}(q_{\phi}(z|x)||p(z))
\end{equation}
where $\beta=1$.  $\beta$-\textbf{VAE} \cite{DBLP:conf/iclr/HigginsMPBGBML17} forces the VAE to learn the disentanglement by setting $\beta>1$. Other unsupervised learning methods \cite{pmlr-v80-kim18b,kumar2018} improve the performance of representation disentanglement by modifying $D_{KL}$, which serves as disentanglement regularization.
Similarly, we add another disentanglement regularization in the loss function by introducing supervision signals using selected pairs of inputs.

\paragraph{Weakly Supervised Swap Variational Autoencoder (\alg):} 
As illustrated in \Cref{fig:algorithm1_2}, the proposed network framework, ~\alg,  consists of two encoders and four decoders, where the parameters within all encoders and decoders are shared respectively. During training, the network is fed a pair of samples $x^{(l)}$ and $x^{(m)}$ and generates representations $z^{(l)}$ and $z^{(m)}$ via encoders $E_1$ and $E_2$ respectively. Decoders $D_1$ and $D_2$ are used to reconstruct  $x^{(l)}$ and $x^{(m)}$ as $x^{(l)}_{rec}$ and $x^{(m)}_{rec}$ respectively; essentially simulating generating input samples $x$ from the distribution $p_{\theta}(x|z)$. Random factors of $z^{(l)}$ and $z^{(m)}$ are swapped to generate two new corresponding latent representations $\hat{z}^{(l)}$ and $\hat{z}^{(m)} $. Decoders $D_3$ and $D_4$ are used to decode the new latent representations $\hat{z}^{(l)}$ and $\hat{z}^{(m)}$ to $\hat{x}^{(l)}_{rec}$ and $\hat{x}^{(m)}_{rec}$.

\begin{figure}
\centering
\includegraphics[width=0.5\linewidth]{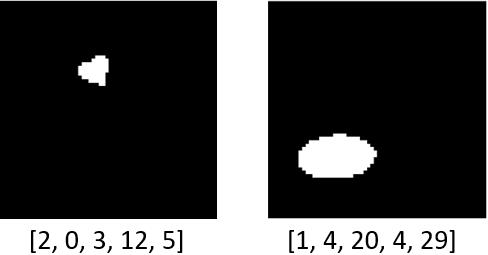}
\caption{Illustration of generating a pair of images of dSprites with different generative factors. The true generative factors values are unavailable during training.}
\label{fig:dsprites_example}
\end{figure}
\paragraph{Detecting distinct generative factors:}
\label{sec:model}
Some latent factors in the representations $z^{(l)}$ and $z^{(m)}$ are fed into a \emph{Detect and Swap} module, which attempts to detect the generative factors of variations that are different between $x^{(l)}$ and $x^{(m)}$. As described in \Cref{eq:pairs}, if the latent representations are fully disentangled, $z^{(l)}$ and $z^{(m)}$ will be same with respect to the dimensions where the same generative factors of $x^{(l)}$ and $x^{(m)}$ are encoded in. $z^{(l)}$ and $z^{(m)}$ will be different in the dimensions where different generative factors of $x^{(l)}$ and $x^{(m)}$ are encoded in. We call the set containing all different underlying factors to be $DF_z$ where $DF_z \subseteq \mathbb{R}^{d_z}$.

\begin{equation}
\begin{split}
\label{eq:pairs}
    p(z_j^{(l)}|\mathbf{x^{(l)}}) &= p(z_j^{(m)}|\mathbf{x^{(m)}}); ~j \notin DF_z \\
    p(z_i^{(l)}|\mathbf{x^{(l)}}) &\neq p(z_i^{(m)}|\mathbf{x^{(m)}}); ~i \in DF_z 
\end{split}
\end{equation}
According to \cite{DBLP:conf/iclr/HigginsMPBGBML17}, the posterior distribution of latent representations is assumed to be Multivariate Gaussian,  $p(z|x)= q_{\theta}(z|x)=\mathcal{N}(\mathbf{\mu},\mathbf{\sigma^2I})$; and the stochastic model becomes differentiable by incorporating reparameterization trick. By using the Multivariate Gaussian assumption and reparamterization trick,  the mutual information between the corresponding dimensions of two latent representation $z^{(l)}$ and $z^{(m)}$ can be directly measured by computing the KL divergence between the distributions of these two latent elements. In practice, by using the posterior distribution of latent representation, we can calculate the KL divergence as shown in \Cref{eq:latentkl}:

\begin{small}
\begin{align}
    \label{eq:latentkl}
    D_{KL}&(q_{\phi}(z_i^{(l)}|x^{(l)})||q_{\phi}(z_i^{(m)}|x^{(m)})) \notag \\
      =&\frac{({\sigma^{(l)}_i})^2 + (\mu^{(l)}_i - \mu^{(m)}_i)^2}{2{(\sigma^{(l)}_i})^2} + log(\frac{\sigma^{(l)}_i}{\sigma^{(m)}_i}) -\frac{1}{2} 
   \end{align}
\end{small}

Instead of using full labels for training, we generate image pairs with at most $k$ distinct factors of variations, and use $k$ as the weak supervision signals for \alg. Since there are at most $k$ distinct generative factors when generating the pair of inputs $x^{(l)}$ and $x^{(m)}$, we expect that there are also at most $k$ latent factors with distinct different values. We assume the top $k$ most distinct underlying factors of variation are the latent factors that produce the highest $k$  KL divergence values for $i=[1, 2,\dots, d_z]$. 
\begin{figure}[]
\centering
\includegraphics[width=0.9\textwidth]{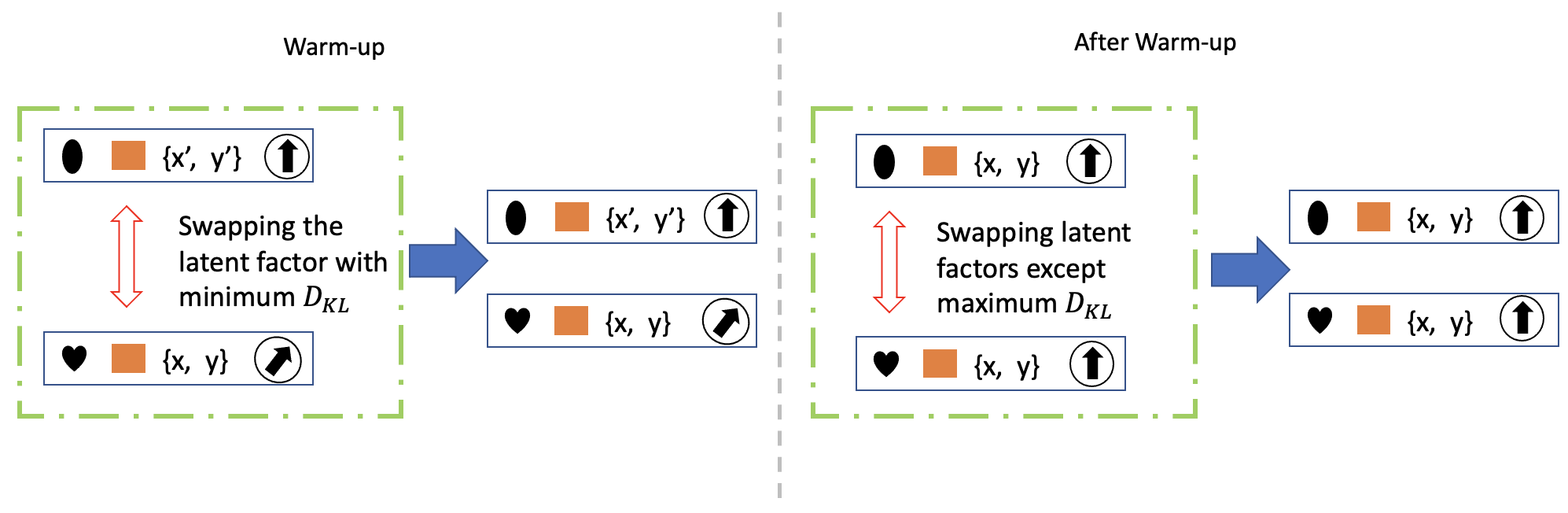}
\caption{Increasing number of latent factors to swap as training progresses. Only one latent factor is swapped during warm up. The number of latent factors to be swapped increases gradually
}
\label{fig:2nd_training_method}
\end{figure}
After detecting the indices of distinct factors in the latent space, we then swap the factors between two latent representations, except the top $k$ most distinct ones. We create two new \emph{swapped} latent representations $\hat{z}^{(l)}$ and $\hat{z}^{(m)}$ using \Cref{eq:exchange}:
\begin{equation}
\begin{split}
    \label{eq:exchange}
    \hat z_i^{(l)} & = z_i^{(m)};  ~  ~~ \hat z_i^{(m)}  = z_i^{(l)};  ~ \forall i \notin DF_z \\
    \hat z_j^{(l)} & = z_j^{(l)};  ~  ~~ \hat z_j^{(m)}  = z_j^{(m)};   ~ \forall j \in DF_z \\ 
\end{split}
\end{equation}

The two swapped representations are then fed into decoders $D_3$ and $D_4$ to produce two reconstructions of input observations $x^{(l)}$ and $x^{(m)}$ as $\hat{x}^{(l)}_{rec}$ and $\hat{x}^{(m)}_{rec}$ respectively. Intuitively, if the network successfully detects all distinct underlying factors of variations, the swapped representations $\hat{z}^{(l)}$ and $\hat{z}^{(m)}$ will be identical to the original representations ${z}^{(l)}$ and ${z}^{(m)}$, 
and therefore the new reconstructed images will be same with the original reconstructions and similar to input observations. This process simultaneously enforces disentangling latent factors and encoding semantic meaning within into these latent factors. Violating either of these two requirements will lead to differences between the new reconstructions and the original reconstructions.  The overall generic loss of the entire network can then be formulated as shown in \Cref{eq:overall_loss}.

\begin{equation}
\begin{split}
    \label{eq:overall_loss}
    L =& L_{VAE}(x^{(l)}_{rec},z^{(l)}) + L_{VAE}(x^{(m)}_{rec},z^{(m)} ) \\
&+ L_g(\hat{x}^{(l)}_{rec}, x^{(l)}_{rec}) + L_g(\hat{x}^{(m)}_{rec}, x^{(m)}_{rec})
\end{split}
\end{equation}
where $L_g$ is the distance function to calculate the difference between $x_{rec}$ and $\hat{x}_{rec}$.  For a detailed study of the behavior of various distance functions, we follow both VAE \cite{kingma2014autoencoding} and VAE-GAN \cite{bib:vaegan}, and introduce two different instantiations of \alg~as follows.

\paragraph{\algmse:} In \algmse~we directly compare the similarity of reconstructions ${x}_{rec}$ and $\hat x_{rec}$ by calculating the mean square error ($\mathbf{MSE}$) loss or binary cross-entropy ($\mathbf{BCE}$) loss, as shown in \Cref{eq:algotithm1mse,eq:algotithm1bce} respectively.
\begin{equation}
\label{eq:algotithm1mse}
  \begin{aligned}
L_{compare\_MSE} =& L_{VAE}(x^{(l)}_{rec},z^{(l)}) +L_{VAE}(x^{(m)}_{rec},z^{(m)} ) \\
    & + \gamma ||\hat{x}^{(l)}_{rec}- x^{(l)}_{rec}||_{2}^2 + \gamma ||\hat{x}^{(m)}_{rec}- x^{(m)}_{rec}||_{2}^2 
 \end{aligned} 
\end{equation}
\begin{figure}[]
\centering
\includegraphics[width=0.9\textwidth]{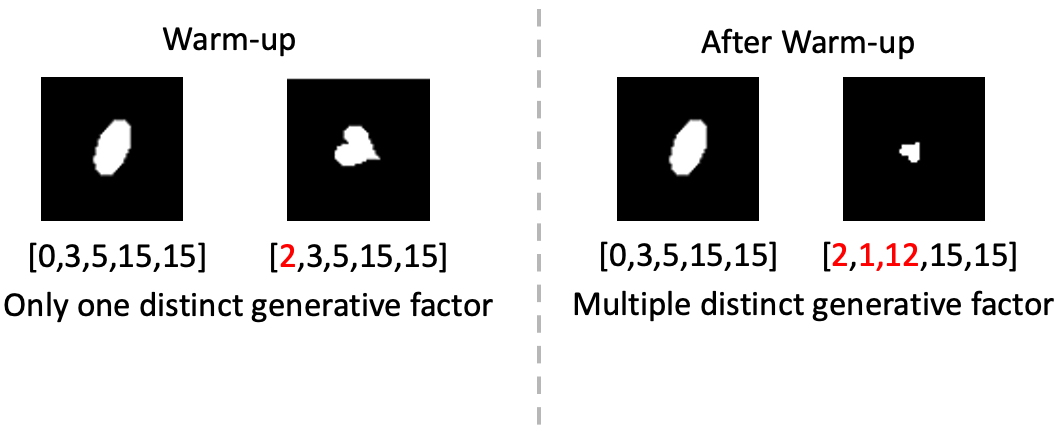}
\caption{Increasing number of distinct generative factors used to produce image pairs as training progresses. }
\label{fig:final_training_method}
\end{figure}
\begin{equation}
\label{eq:algotithm1bce}
  \begin{aligned}
L_{compare\_BCE} =& L_{VAE}(x^{(l)}_{rec},z^{(l)}) +L_{VAE}(x^{(m)}_{rec},z^{(m)} ) \\
    & + \gamma \mathbf{BCE}(\hat{x}^{(l)}_{rec}, x^{(l)}_{rec}) +\gamma \mathbf{BCE}(\hat{x}^{(m)}_{rec}, x^{(m)}_{rec}) 
 \end{aligned} 
\end{equation}

\paragraph{\alggan:} In \alggan~we use a discriminator to measure the similarity between reconstruction pairs. During training, the VAE tries to minimize the distance between $x_{rec}$ and $\hat{x}_{rec}$. The discriminator tries to measure the differences between two input images.  We use binary cross-entropy loss to train the VAE and discriminator. Image labels are set to \emph{one} when training VAE and labels are set to \emph{zero} when training discriminator.  In \alggan, the loss function to train VAE is expressed in \Cref{eq:algorithm2} and the loss function to train the discriminator is expressed in \Cref{eq:discriminator}.

\begin{equation}
\begin{aligned}
    \label{eq:algorithm2}
    L_{GAN} =& L_{VAE}(x^{(l)}_{rec},z^{(l)}) +L_{VAE}(x^{(m)}_{rec},z^{(m)} ) \\ %
    & +  \gamma \mathbf{BCE}(C_{w}\Big[\hat{x}^{(l)}_{rec},x^{(l)}_{rec}\Big],\mathbf{1}) + \gamma \mathbf{BCE}(C_{w}\Big[\hat{x}^{(m)}_{rec},x^{(m)}_{rec}\Big],\mathbf{1})
\end{aligned}
\end{equation}

\begin{equation}
\begin{aligned}
\label{eq:discriminator}
    L_{disc}  =& \mathbf{BCE}(C_{w}\Big[\hat{x}^{(l)}_{rec},x^{(l)}_{rec}\Big],\mathbf{0}) + \mathbf{BCE}(C_{w}\Big[\hat{x}^{(m)}_{rec},x^{(m)}_{rec}\Big],\mathbf{0})
\end{aligned}
\end{equation}

The overall architecture of \alg~is shown in \Cref{fig:algorithm1_2}. The difference between these two algorithms is the measurement criterion of comparing the similarity between the reconstruction output $x_{recon}$ from original the latent representations and the new reconstruction output $\hat{x}_{recon}$ from the new latent representations.

\paragraph{Training using maximum number of different generative factors:}
\label{paragraph:pairs}
As previously discussed, Locatello \etal~\cite{locatello2019challenging} argue that disentanglement of the underlying generative factors of variations is infeasible without any supervision signal. In this work, we use pairs of input $x^{(l)}$ and $x^{(m)}$  as weak supervision signals to learn disentangled representations.

\begin{table*}[]
\caption{Disentanglement metrics on dSprites and 3dShapes; \red{\textbf{Bold, Red}}: best result, \textbf{Bold, Black}: second best result; \alg(m) is trained only using the maximum number of generative factors, and \alg(e) is trained with knowledge of the exact number of different generative factors; Baseline results are generated by us with official public implementation.}
\label{table:dsprites_shapes3d}
\centering
\tiny
\renewcommand{\arraystretch}{1.2}
{
\begin{adjustbox}{width=1\textwidth}
\begin{tabular}{cllllllllll}
\hlineB{3}
\multicolumn{1}{c|}{\multirow{2}{*}{\textbf{Models}}} & \multicolumn{5}{c|}{\textbf{dSprites}}                                                                                                                                         & \multicolumn{5}{c}{\textbf{3dshapes}}                                                                                                                     \\ \cline{2-11} 
\multicolumn{1}{c|}{}                                 & \multicolumn{1}{c}{\textbf{MIG}} & \multicolumn{1}{c}{\textbf{SAP}} & \multicolumn{1}{c}{\textbf{IRS}} & \multicolumn{1}{c}{\textbf{FVAE}} & \multicolumn{1}{l|}{\textbf{DCI}} & \multicolumn{1}{c}{\textbf{MIG}} & \multicolumn{1}{c}{\textbf{SAP}} & \multicolumn{1}{c}{\textbf{IRS}} & \multicolumn{1}{c}{\textbf{FVAE}} & \textbf{DCI} \\ \hline
\multicolumn{11}{c}{\textbf{Unsupervised Disentanglement Learning}}                                                                                                                                                                                                                                                                                                                                \\ \hline
\multicolumn{1}{c|}{betaVAE}                          & \multicolumn{1}{c}{0.1115}       & \multicolumn{1}{c}{0.0373}       & \multicolumn{1}{c}{0.5520}       & \multicolumn{1}{c}{0.7886}        & \multicolumn{1}{l|}{0.3263}       & \multicolumn{1}{c}{0.1868}       & 0.0643                           & 0.4731                           & 0.8488                            & 0.2460       \\
\multicolumn{1}{c|}{AnnealedVAE}                      & \multicolumn{1}{c}{0.1253}       & \multicolumn{1}{c}{0.0422}       & \multicolumn{1}{c}{0.5592}       & \multicolumn{1}{c}{0.8228}        & \multicolumn{1}{l|}{0.3715}       & \multicolumn{1}{c}{0.2350}       & 0.0867                           & 0.5450                           & 0.8651                            & 0.3428       \\
\multicolumn{1}{c|}{FactorVAE}                        & \multicolumn{1}{c}{0.1308}       & \multicolumn{1}{c}{0.0238}       & \multicolumn{1}{c}{0.5634}       & \multicolumn{1}{c}{0.7503}        & \multicolumn{1}{l|}{0.1914}       & \multicolumn{1}{c}{0.2265}       & 0.0437                           & 0.6298                           & 0.7876                            & 0.3031       \\
\multicolumn{1}{c|}{DIP-VAE-I}                        & 0.0667                           & 0.0247                           & 0.4497                           & 0.6897                            & \multicolumn{1}{l|}{0.1661}       & 0.1438                           & 0.0273                           & 0.5010                           & 0.7671                            & 0.1372       \\
\multicolumn{1}{c|}{DIP-VAE-II}                       & 0.0212                           & 0.0587                           & 0.5434                           & 0.6354                            & \multicolumn{1}{l|}{0.0997}       & 0.1372                           & 0.0204                           & 0.4237                           & 0.7416                            &              \\
\multicolumn{1}{c|}{BetaTCVAE}                        & 0.2125                           & 0.0582                           & 0.5437                           & 0.8414                            & \multicolumn{1}{l|}{0.3295}       & 0.3644                           & 0.0955                           & 0.5942                           & 0.9627                            & 0.6004       \\ \hline
\multicolumn{11}{c}{\textbf{Weakly/Semi-Supervised Disentanglement Learning}}                                                                                                                                                                                                                                                                                                                      \\ \hline
\multicolumn{1}{c|}{DSD}                              & 0.3937                           & 0.0771                           & 0.6327                           & \red{\textbf{0.9121}}                            & \multicolumn{1}{l|}{\textbf{0.6015}}       & 0.6343                           & \textbf{0.1518}                           & \textbf{0.7623}                           & \textbf{0.9968}                            & 0.9019       \\
\multicolumn{1}{c|}{Ada-ML-VAE}                       & 0.1150                           & 0.0366                           & 0.5712                           & 0.7010                            & \multicolumn{1}{l|}{0.2940}       & 0.5092                           & 0.1273                           & 0.6203                           & 0.9956                            & \textbf{0.9400}       \\
\multicolumn{1}{c|}{Ada-GVAE}                         & 0.2664                           & 0.0735                           & 0.5927                           & 0.8472                            & \multicolumn{1}{l|}{0.4790}       & 0.5607                           & 0.1502                           & 0.7076                           & 0.9965                            & \red{\textbf{0.9459}}       \\ \hdashline
\multicolumn{1}{c|}{\alg(m)}                    & \textbf{0.4228}             & \textbf{0.0780}                           & \textbf{0.6572}                           & 0.8571                            & \multicolumn{1}{l|}{0.5994}       & \textbf{0.6353}                           & 0.1506                           & 0.7316                           & 0.9958                            & 0.9030       \\
\multicolumn{1}{c|}{\alg(e)}                    & \red{\textbf{0.4637}}                           & \red{\textbf{0.1077}}                           & \red{\textbf{0.6774}}                           & \textbf{0.8913}                            & \multicolumn{1}{l|}{\red{\textbf{0.6852}}}       & \red{\textbf{0.7121}}                           & \red{\textbf{0.1564}}                           & \red{\textbf{0.7837}}                           & \red{\textbf{0.9978}}                            & 0.9198       \\ \hlineB{3}
\end{tabular}
\end{adjustbox}
}

\end{table*}

As  illustrated in \Cref{fig:dsprites_example}, the pair of training samples ($x^{(l)}$, $x^{(m)}$) can be generated as follow: we first randomly select a vector $V^{(l)} = [v_1,v_2, ..., v_n] $ to generate observation $x^{(l)}=g(V^{(l)})$, where $g(V)$ is the observation generating function. Then, we randomly change the value of at most $k$ elements in $V^{(l)}$ to form a new vector $V^{(m)}$. Finally, we generate the $x^{(m)}=g(V^{(m)})$. 
During training, true indices of different generative factors and true value of generative factors are not provided, where the only information provided to the model is the maximum number of changed factors $k$, as mentioned earlier.

During training, there are more than one factors of variants that need to be swapped. However, in the early training stage, since the model is not well trained yet, exchanging a large number of factors in the latent representations tends to harm model performance, which is discussed in \Cref{sec:ablation}. Thus, during \emph{warm-up}, we only swap the factors of latent representation where the model is highly confident. As training progresses, we increase the difficulty by increasing the number of latent representation factors to swap. We show this procedure in \Cref{fig:2nd_training_method} and call this strategy as ISF. We denote \alg~trained under the supervision of maximum number of changed factors as \alg(m).

\begin{table*}[]
\caption{Disentanglement metrics on MPI3D-toy, MPI3D-realistic, and MPI3D-real; \red{\textbf{Bold, Red}}: best result, \textbf{Bold, Black}: second best result; \alg(m) is trained only using the maximum number of generative factors, and \alg(e) is trained with knowledge of the exact number of different generative factors; Baseline results are generated by us with official public implementation.}
\label{table:mpi3d}
\renewcommand{\arraystretch}{1.3}
{
\begin{adjustbox}{width=\textwidth}
\begin{tabular}{clllllllllllllll}
\hlineB{3}
\multicolumn{1}{c|}{\multirow{2}{*}{\textbf{Models}}} & \multicolumn{5}{c|}{\textbf{MPI3d-toy}}                                                                                                                                        & \multicolumn{5}{c|}{\textbf{MPI3d-realistic}}                                                                                                                                  & \multicolumn{5}{c}{\textbf{MPI3d-real}}                                                                                                                   \\ \cline{2-16} 
\multicolumn{1}{c|}{}                                 & \multicolumn{1}{c}{\textbf{MIG}} & \multicolumn{1}{c}{\textbf{SAP}} & \multicolumn{1}{c}{\textbf{IRS}} & \multicolumn{1}{c}{\textbf{FVAE}} & \multicolumn{1}{l|}{\textbf{DCI}} & \multicolumn{1}{c}{\textbf{MIG}} & \multicolumn{1}{c}{\textbf{SAP}} & \multicolumn{1}{c}{\textbf{IRS}} & \multicolumn{1}{c}{\textbf{FVAE}} & \multicolumn{1}{l|}{\textbf{DCI}} & \multicolumn{1}{c}{\textbf{MIG}} & \multicolumn{1}{c}{\textbf{SAP}} & \multicolumn{1}{c}{\textbf{IRS}} & \multicolumn{1}{c}{\textbf{FVAE}} & \textbf{DCI} \\ \hline
\multicolumn{16}{c}{\textbf{Unsupervised Disentanglement Learning}}                                                                                                                                                                                                                                                                                                                                                                                                                                                                                                                 \\ \hline
\multicolumn{1}{c|}{betaVAE}                          & \multicolumn{1}{c}{0.132}        & \multicolumn{1}{c}{0.061}        & \multicolumn{1}{c}{0.584}        & \multicolumn{1}{c}{0.353}         & \multicolumn{1}{l|}{0.274}        & \multicolumn{1}{c}{0.150}        & 0.138                            & 0.574                            & 0.350                             & \multicolumn{1}{l|}{0.361}        & 0.137                            & 0.071                            & 0.579                            & 0.368                             & 0.367        \\
\multicolumn{1}{c|}{AnnealedVAE}                      & \multicolumn{1}{c}{0.155}        & \multicolumn{1}{c}{0.107}        & \multicolumn{1}{c}{0.553}        & \multicolumn{1}{c}{0.411}         & \multicolumn{1}{l|}{0.360}        & \multicolumn{1}{c}{0.109}        & 0.114                            & 0.531                            & 0.466                             & \multicolumn{1}{l|}{0.363}        & 0.099                            & 0.038                            & 0.490                            & 0.396                             & 0.229        \\
\multicolumn{1}{c|}{FactorVAE}                        & \multicolumn{1}{c}{0.183}        & \multicolumn{1}{c}{0.070}        & \multicolumn{1}{c}{0.512}        & \multicolumn{1}{c}{0.398}         & \multicolumn{1}{l|}{0.273}        & \multicolumn{1}{c}{0.136}        & 0.069                            & 0.560                            & 0.386                             & \multicolumn{1}{l|}{0.215}        & 0.093                            & 0.031                            & 0.529                            & 0.391                             & 0.192        \\
\multicolumn{1}{c|}{DIP-VAE-I}                        & 0.156                            & 0.085                            & 0.484                            & 0.517                             & \multicolumn{1}{l|}{0.284}        & \multicolumn{1}{c}{0.167}        & 0.134                            & 0.525                            & 0.566                             & \multicolumn{1}{l|}{0.337}        & 0.130                            & 0.074                            & 0.509                            & 0.533                             & 0.264        \\
\multicolumn{1}{c|}{DIP-VAE-II}                       & 0.061                            & 0.027                            & 0.412                            & 0.487                             & \multicolumn{1}{l|}{0.163}        & 0.032                            & 0.017                            & 0.432                            & 0.393                             & \multicolumn{1}{l|}{0.149}        & 0.131                            & 0.062                            & 0.509                            & 0.544                             & 0.244        \\
\multicolumn{1}{c|}{BetaTCVAE}                        & 0.173                            & 0.084                            & \red{\textbf{0.638}}                            & 0.355                             & \multicolumn{1}{l|}{0.342}        & 0.165                            & 0.046                            & 0.571                            & 0.366                             & \multicolumn{1}{l|}{0.244}        & 0.181                            & 0.146                            & \red{\textbf{0.636}}                            & 0.431                             & 0.344        \\ \hline
\multicolumn{16}{c}{\textbf{Weakly/Semi-Supervised Disentanglement Learning}}                                                                                                                                                                                                                                                                                                                                                                                                                                                                                          \\ \hline
\multicolumn{1}{c|}{DSD}                              & \textbf{0.399}                            & 0.193                            & 0.612                            & \textbf{0.510}                             & \multicolumn{1}{l|}{0.445}        & 0.402                            & 0.199                            & 0.602                            & \red{\textbf{0.621}}                             & \multicolumn{1}{l|}{0.512}        & 0.353                            & \textbf{0.222}                            & 0.601                            & \red{\textbf{0.603}}                             & 0.522        \\
\multicolumn{1}{c|}{Ada-ML-VAE}                       & 0.293                            & 0.093                            & 0.520                            & 0.439                             & \multicolumn{1}{l|}{0.392}        & 0.283                            & 0.131                            & 0.582                            & 0.483                             & \multicolumn{1}{l|}{0.270}        & 0.240                            & 0.074                            & 0.576                            & 0.476                             & 0.285        \\
\multicolumn{1}{c|}{Ada-GVAE}                         & 0.347           & \red{\textbf{0.238}}                            & 0.613                            & 0.501                             & \multicolumn{1}{l|}{0.427}        & 0.339                            & \red{\textbf{0.227}}                            & 0.609                            & 0.592                             & \multicolumn{1}{l|}{0.479}        & 0.264                            & 0.215                            & 0.602                            & \textbf{0.601}                             & 0.401        \\\hdashline
\multicolumn{1}{l|}{~~\alg(m)}                    & 0.394                            & 0.210                            & 0.592                            & 0.506                             & \multicolumn{1}{l|}{\textbf{0.452}}        & 0.401                            & 0.201                            & 0.602                            & 0.591                             & \multicolumn{1}{l|}{0.532}        & \textbf{0.360}                            & 0.216                            & 0.593                            & 0.562                             & \textbf{0.542}        \\ 
\multicolumn{1}{l|}{~~\alg(e)}                  & \red{\textbf{0.467}}      & \textbf{0.213}                            & \textbf{0.614}                            & \red{\textbf{0.520}}                             & \multicolumn{1}{l|}{\red{\textbf{0.554}}}        & \red{\textbf{0.452}}    & \textbf{0.202}                            & \red{\textbf{0.614}}                            & \textbf{0.596}                             & \multicolumn{1}{l|}{\red{\textbf{0.574}}}        & \red{\textbf{0.484}}                            & \red{\textbf{0.225}}                            & \textbf{0.617}                           & 0.565                             & \red{\textbf{0.566}}        \\ \hlineB{3}
\end{tabular}
\end{adjustbox}
}

\end{table*}
\paragraph{Training using exact number of different generative factors:}
As previously discussed, \alg(m) is limited to merely know the maximum number $k$ of different generative factors. By slightly increasing the level of supervision strength, where the exact number of different generative factors of the pair of inputs are known, the performance of the model is further boosted though the true indices and values of different generative factors are still inaccessible for the model. We denote \alg~trained under supervision of exact different generative factors as  \alg(e).   During training, we set the number of different generative factors to one and progressively increasing until it reaches the maximum number. This strategy is illustrated in \Cref{fig:final_training_method} and we call this strategy as IGF. The importance of this training strategies is studied in section \Cref{sec:ablation}.

\section{Experimental Evaluation}

\label{sec:experiment}
\subsection{Benchmarks, Baseline Methods and Evaluation Metrics}
\label{dataset}
We use following five different datasets where the images are annotated with different underlying factor of variations.

\textbf{dSprites} \cite{DBLP:conf/iclr/HigginsMPBGBML17} contains 73,728 binary $64\times64$ images generated by 6 generative factors.  

\textbf{3dShapes} \cite{3dshapes18}  generated by 6 generative factors: floor hue, wall hue, object hue, scale, shape and orientation. The total dataset contains 480,000 RGB $64\times64\times3$ images.  

\textbf{MPI3D} \cite{gondal2019transfer} contains 3 different datsets: MPI3D-real, MPI3D-toy and MPI3D-realistic. MPI3D-real is a real world dataset controlled by 7 generative factors: object color, object shape, object size, camera height, background color, horizontal axis and vertical axis.   MPI3D-toy,  MPI3D-realistic are synthetic versions of MPI3D-real with two levels of realism. Each of the three datasets contains 1,036,800 RGB $512\times512\times3$ images.

SOTA model used for comparisons are: (1) $\beta$-VAE \cite{DBLP:conf/iclr/HigginsMPBGBML17}, (2) AnnealedVAE \cite{burgess2018understanding}, (3) FactorVAE \cite{pmlr-v80-kim18b}, (4) DIP-VAE-I \cite{kumar2018}, (5) DIP-VAE-II \cite{kumar2018}, (6) $\beta$-TCVAE \cite{chen2019isolating},  (7) DSD \cite{NEURIPS2018_fdf1bc56}, (8) Ada-ML-VAE \cite{Locatello2020Disentangling} and (9) Ada-VAE \cite{Locatello2020Disentangling}.

\begin{figure}[]

\centering
\begin{tabular}{cccc}
\includegraphics[width=0.3\textwidth]{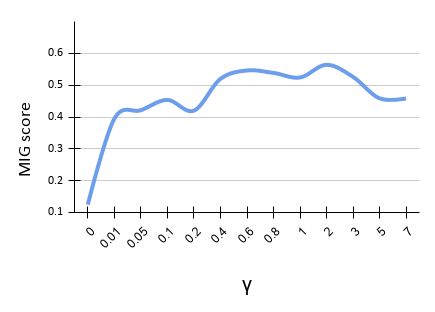} &
\includegraphics[width=0.3\textwidth]{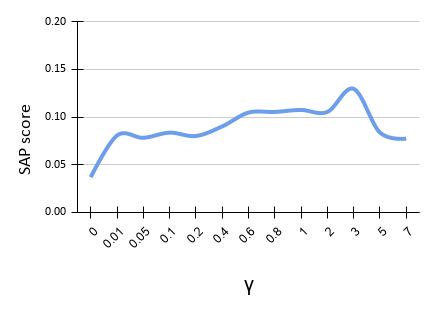} \\
\textbf{(a)}  & \textbf{(b)} \\[6pt]
\end{tabular}
\begin{tabular}{cccc}
\includegraphics[width=0.3\textwidth]{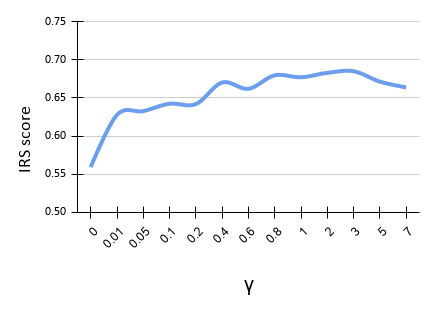} &
\includegraphics[width=0.3\textwidth]{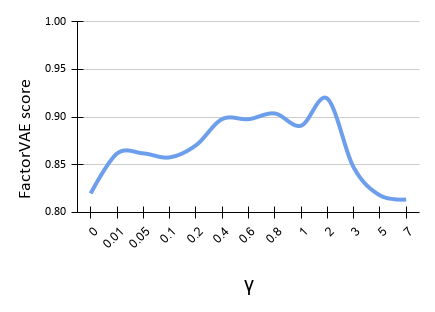} &
\includegraphics[width=0.3\textwidth]{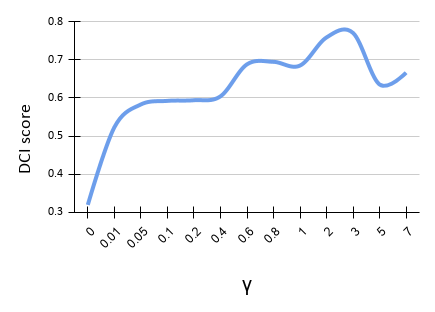} \\
 \textbf{(c)} & \textbf{(e)}  & \textbf{(f)}   \\[6pt]
\end{tabular}
\centering
\caption{Disentanglement metrics with varying $\gamma$ on dSprites.
\textbf{(a)} Mutual Information Gap (MIG)
\textbf{(b)} Separated Attribute Predictability (SAP) 
\textbf{(c)} Interventional Robustness Score (IRS)
\textbf{(d)} FactorVAE score (FVAE)
\textbf{(e)} DCI-Disentanglement (DCI)
}
\label{fig:hyper-parameter}
\end{figure}
Following metrics are used to evaluate the performance of \alg. All metrics are range from $0$ to $1$, where the score of $1$ indicates the latent factors are fully disentangled.

\begin{itemize}
    \item \textbf{Mutual Information Gap (MIG)} \cite{chen2019isolating} calculates the difference between the top two highest mutual information between latent and generative components.  
    \item \textbf{Separated Attribute Predictability (SAP)} \cite{kumar2018} calculates the average perdition error difference between the top two most predictive latent components. 
    \item \textbf{Interventional Robustness Score (IRS)} \cite{suter2019robustly} assesses the degree of dependency between a latent factor and a generative factor, regardless of additional generative factors. 
    \item \textbf{FactorVAE (FVAE) score} \cite{pmlr-v80-kim18b} predicts the index of a fixed generating factor using a majority vote classifier, and the accuracy is the final score value. 
    \item \textbf{DCI-Disentanglement (DCI)} \cite{eastwood2018a} computes the entropy of the distribution by normalizing across each dimension of the learned representation in order to predict the value of a generative component. 
\end{itemize}

\subsection{Quantitative Results}
\label{sec:quanity}
\Cref{table:dsprites_shapes3d,table:mpi3d} show disentanglement metrics results tested on dSprites, 3dshapes, MPI3D-toy, MPI3D-realistic and MPI3D-real respectively.
By observing the results, \alg~outperforms baselines methods in most cases, where \alg(e)~consistently outperforms \alg(m). Mean, median, and variance of all disentanglement metrics scores of all models tested on all datasets are shown Appendix.

\subsection{Ablation Study}
\label{sec:ablation}
\paragraph{Effectiveness Of New Regularization:}
To prove the effectiveness of using the proposed regularization as discussed in \Cref{eq:overall_loss,eq:algotithm1mse,eq:algotithm1bce,eq:algorithm2}, we evaluated \alg(m) on dSprites using different values of $\gamma$. We start experimenting by setting $\gamma=0$, then slowly increase the value of $\gamma$. We show the change of disentanglement metrics MIG and DCI scores in \Cref{fig:hyper-parameter}. We can observe a great improvement by utilizing swapping concept even if $\gamma$ is very small. Other metrics results are included in supplementary materials.

\begin{table}[]
\caption{Disentanglement metrics of \alg(m) with different training strategies applied to dSpirits dataset}
\label{table:strategy_algorithm1}
\centering
\small
\renewcommand{\arraystretch}{1.3}
{
\begin{adjustbox}{width=0.5\textwidth}
\begin{tabular}{M{0.75cm} M{0.75cm}| M{0.75cm} M{0.75cm} M{0.75cm} M{0.75cm} M{0.75cm}}
\hlineB{2}
\textbf{ISF} & \textbf{IGF} & \textbf{MIG}     & \textbf{SAP}    & \textbf{IRS}    & \textbf{FVAE} & \textbf{DCI}   \\ \hline
& & 0.232 & 0.048  & 0.617  & 0.790   & 0.347 \\ 
& \checkmark   & 0.439 & 0.081 & 0.652 & 0.872 & 0.610 \\ 
\checkmark & \checkmark & \textbf{0.525} & \textbf{0.108} & \textbf{0.677} & \textbf{0.891}  & \textbf{0.685} \\ \hlineB{2}
\end{tabular}
\end{adjustbox}
}
\end{table}

\paragraph{Effectiveness Of Different Training Strategies:}

As we have discussed in \Cref{sec:methods}, we propose two training strategies (gradually increasing the number of swapped latent factors (ISF) shown in \Cref{fig:2nd_training_method} and gradually increasing the number of different generative factors (IGF) shown in \Cref{fig:final_training_method} to help learning representation disentanglement. We study the importance of these strategies by comparing the results of three different situations: (1) No strategy is used; (2) Only IGF strategy is used ; (3) Both IGF and ISF strategies are used. The result of disentanglement performance evaluated on dSprites with different strategies are shown in \Cref{table:strategy_algorithm1}. Compared to none of the proposed training strategies is used, we can observe significant  performance improvement by adopting IGF strategy. Furthermore, there is additional performance improvement by utilizing both the ISF and IGF strategies.

\section{Conclusion}

We introduce \alg: a novel weakly-supervised representation disentanglement method. The supervision signals are introduced by utilizing pairs of training observations where the number of different generative factors are controlled. \alg~uses the maximum or 
exact number of different generative factors as an instruction to swap the latent factors estimated by the encoder. Further, the comparison between the reconstruction from the original latent representation and the reconstruction from the new latent representation serves as new disentanglement regularization. Experimental evaluation demonstrates that our approach significantly outperforms SOTAs both qualitatively and quantitatively.

\noindent\textbf{Acknowledgement:} This material is based on research sponsored by Air Force Research Laboratory under agreement number FA8750-19-1-1000. The views and conclusions contained herein are those of the authors and should not be interpreted as necessarily representing the official policies or endorsements, either expressed or implied, of Air Force Research Laboratory or the US, Government.

\bibliographystyle{splncs04}
\bibliography{eccv_workshop}

\clearpage
\section{Appendix}
\subsection{Box plot of All metrics}
\begin{figure*}[]

\centering
\begin{tabular}{cccc}
\includegraphics[width=0.3\textwidth,height=0.29\textwidth]{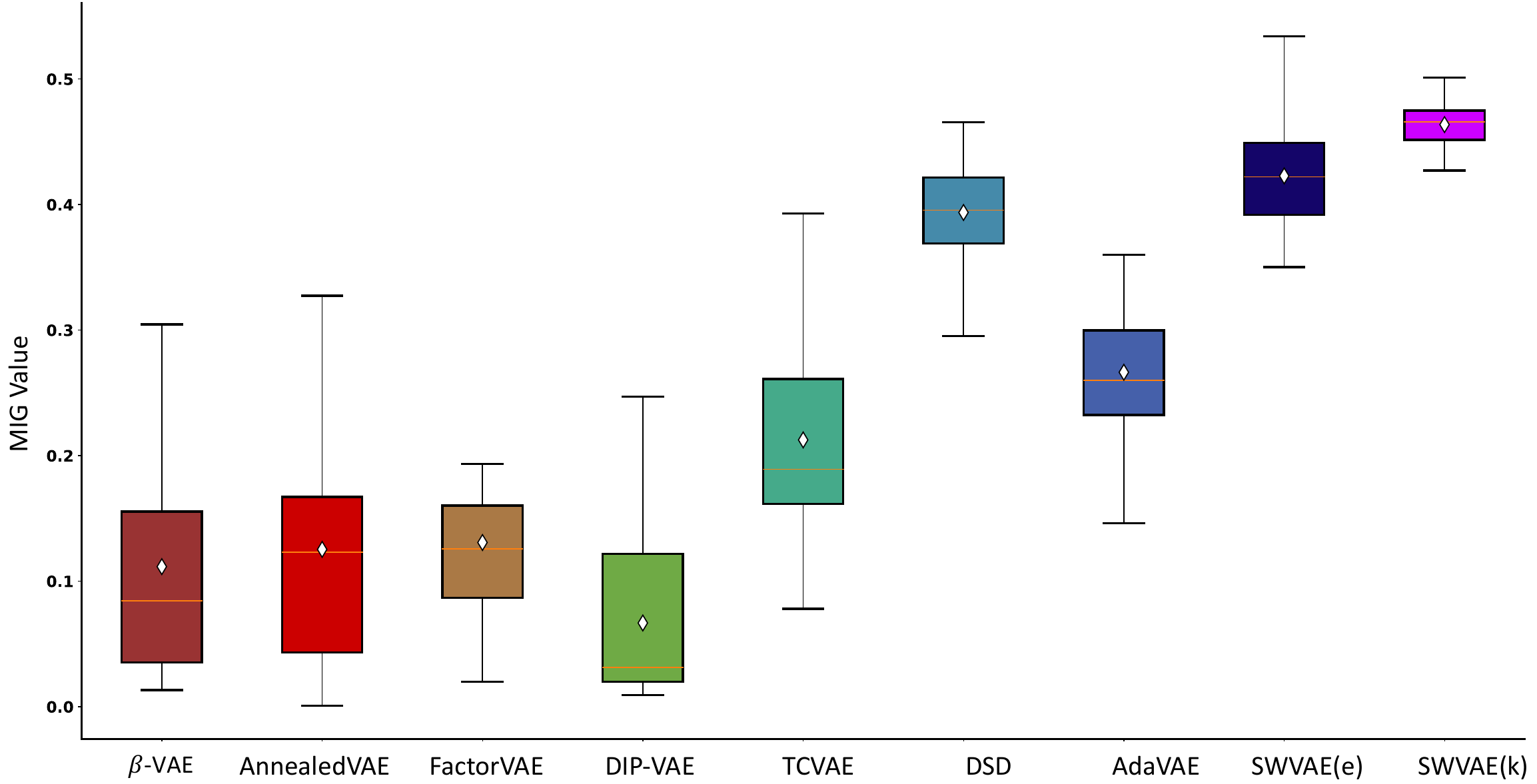} &
\includegraphics[width=0.3\textwidth,height=0.29\textwidth]{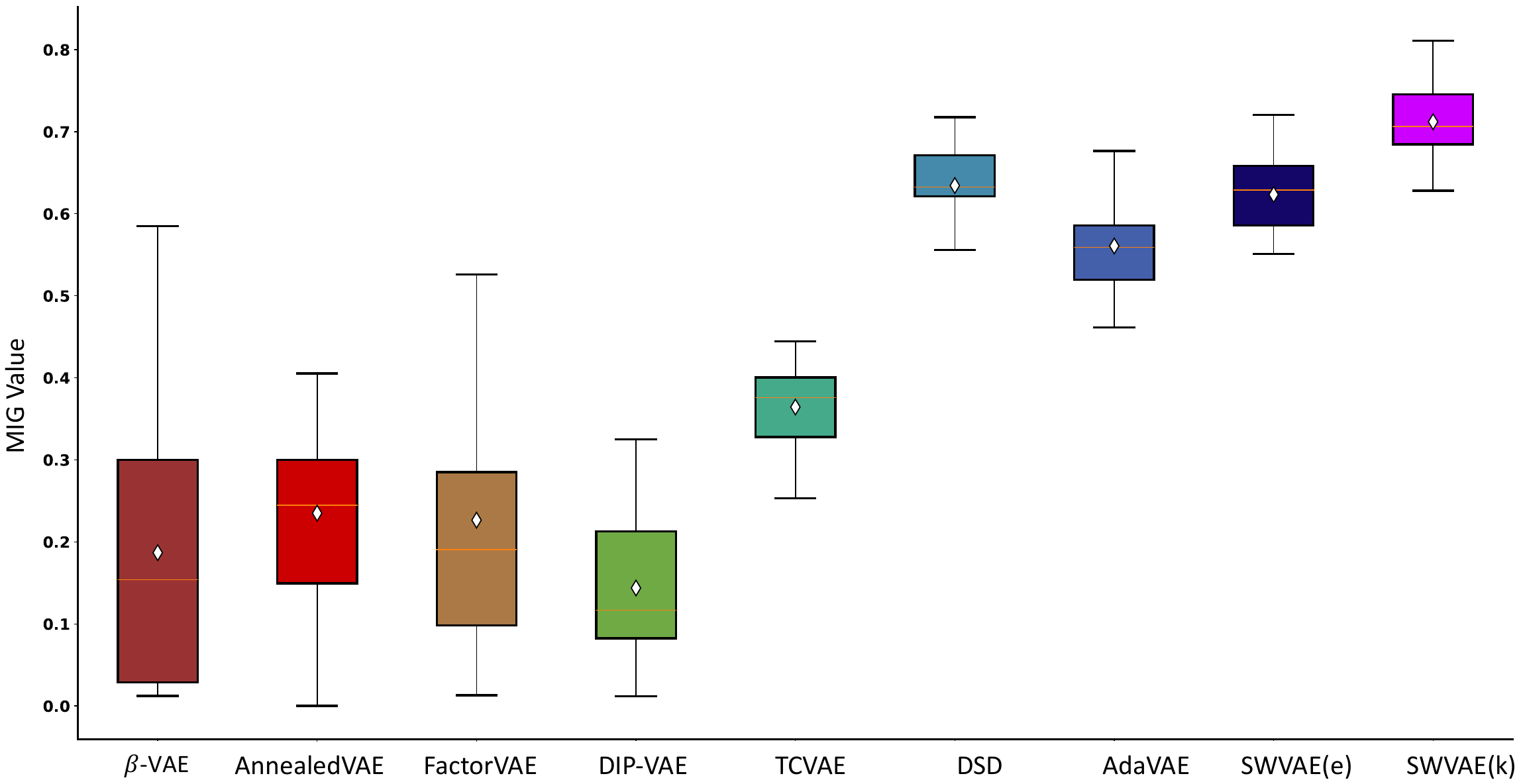} &
\includegraphics[width=0.3\textwidth,height=0.29\textwidth]{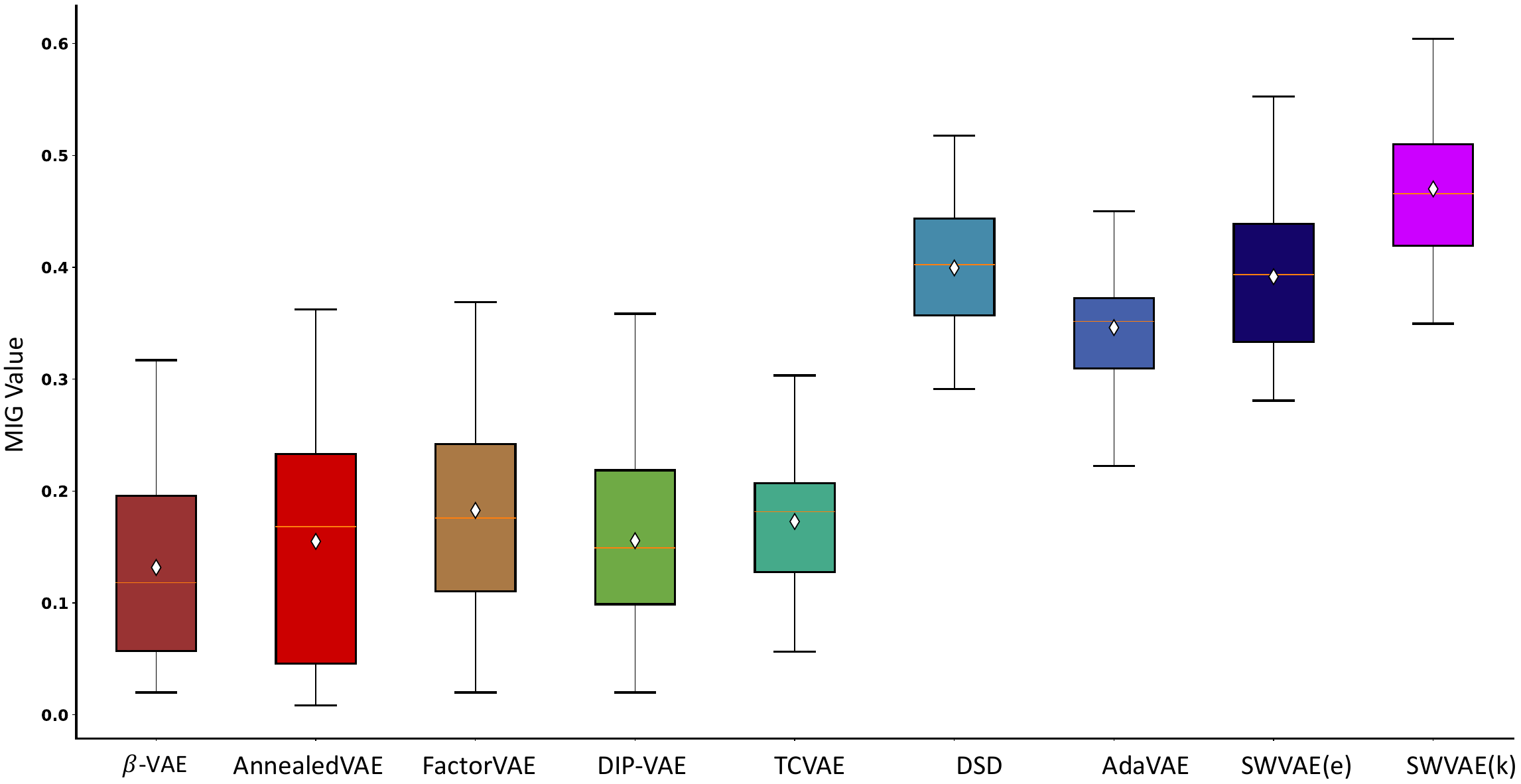} \\
\textbf{(a)}  & \textbf{(b)} & \textbf{(c)}  \\[6pt]
\end{tabular}
\begin{tabular}{cccc}
\includegraphics[width=0.3\textwidth,height=0.29\textwidth]{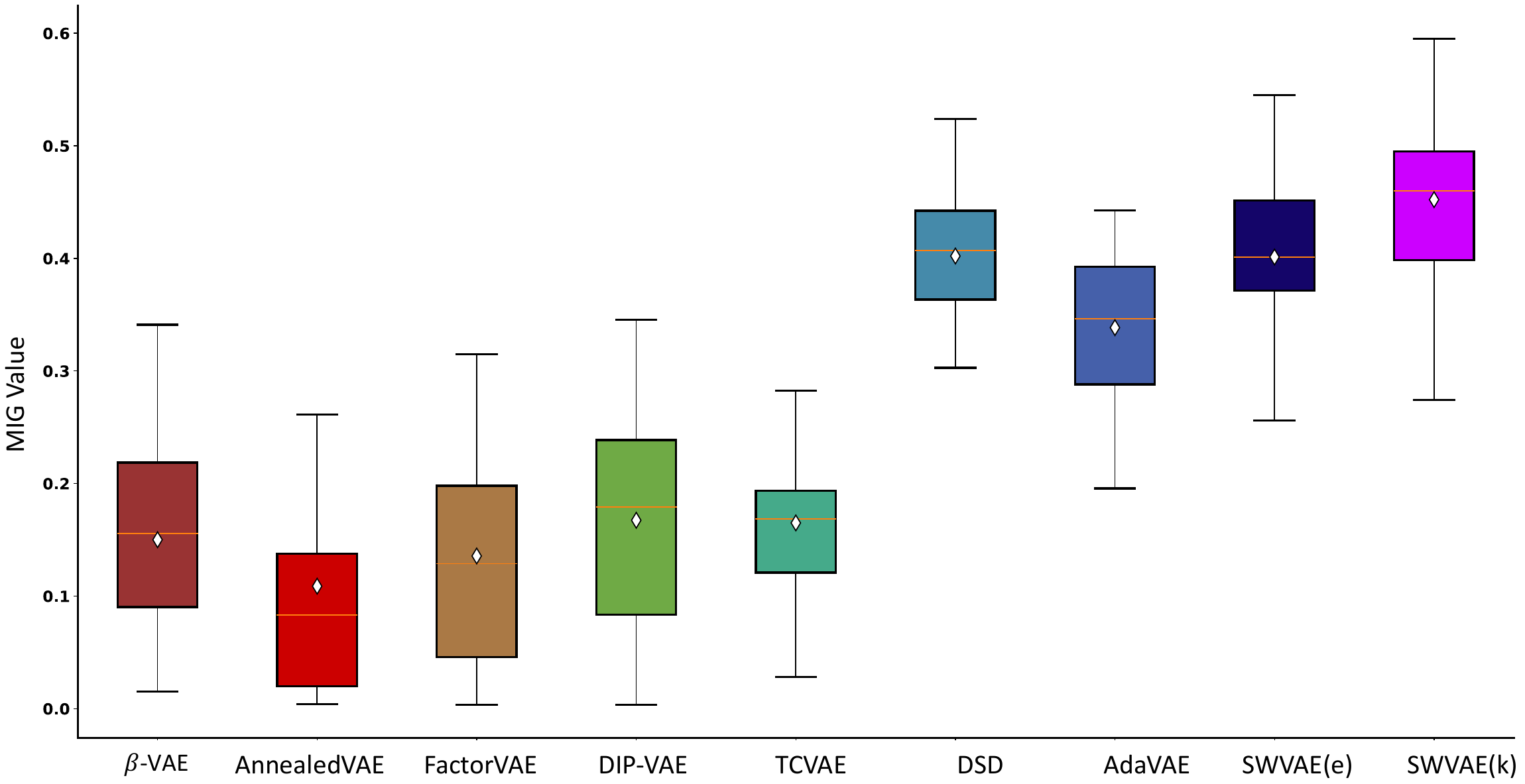} &
\includegraphics[width=0.3\textwidth,height=0.29\textwidth]{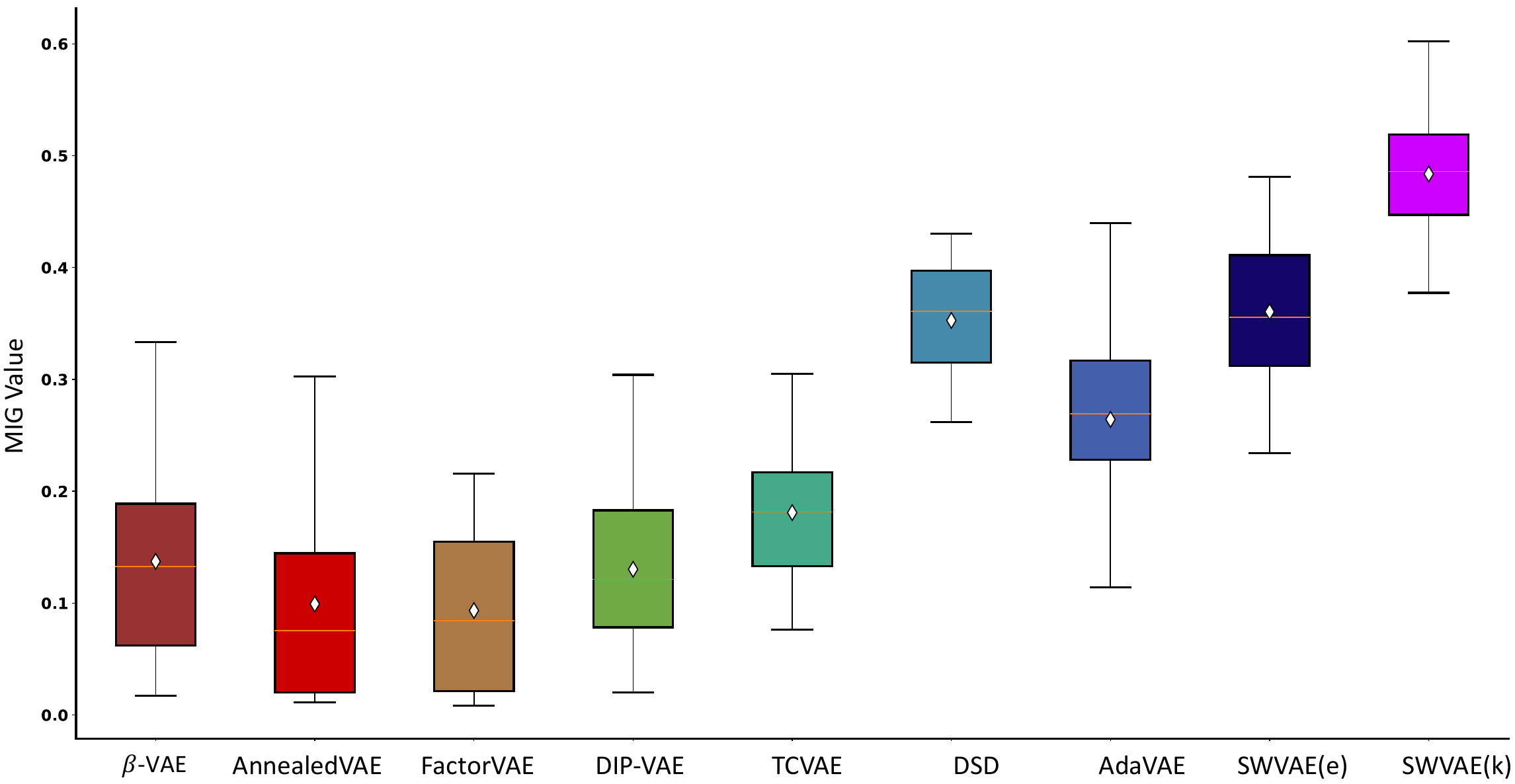} \\
\textbf{(d)}  & \textbf{(e)}   \\[6pt]
\end{tabular}
\centering
\caption{Box plot of MIG scores among models tested on different datasets. 
\textbf{(a)} dSprites 
\textbf{(b)} 3dSahpes
\textbf{(c)} MPI3D-toy
\textbf{(d)} MPI3D-realistic
\textbf{(e)} MPI3D-real
}
\label{fig:MIG}
\end{figure*}

\begin{figure*}[]

\centering
\begin{tabular}{cccc}
\includegraphics[width=0.3\textwidth,height=0.29\textwidth]{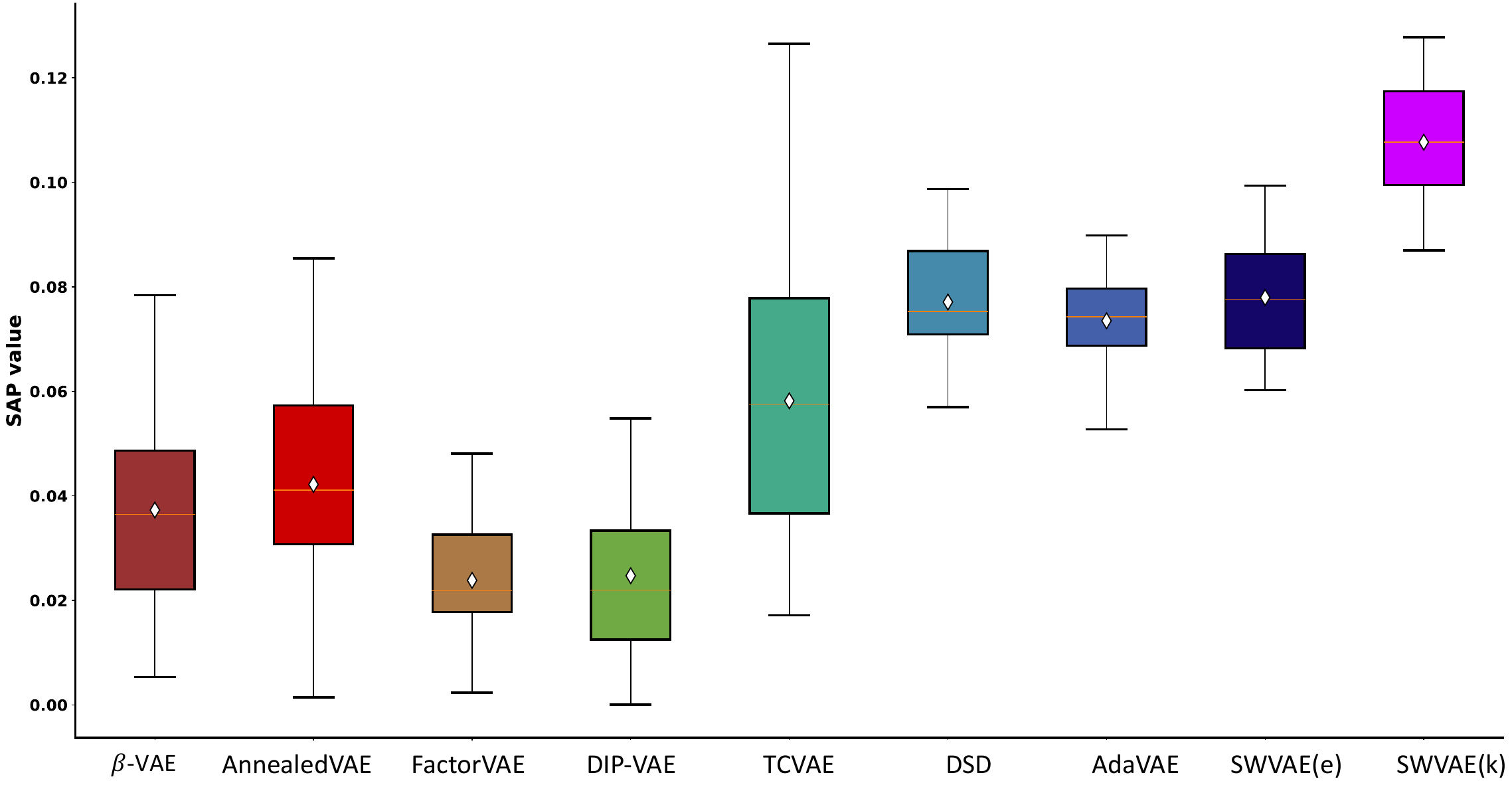} &
\includegraphics[width=0.3\textwidth,height=0.29\textwidth]{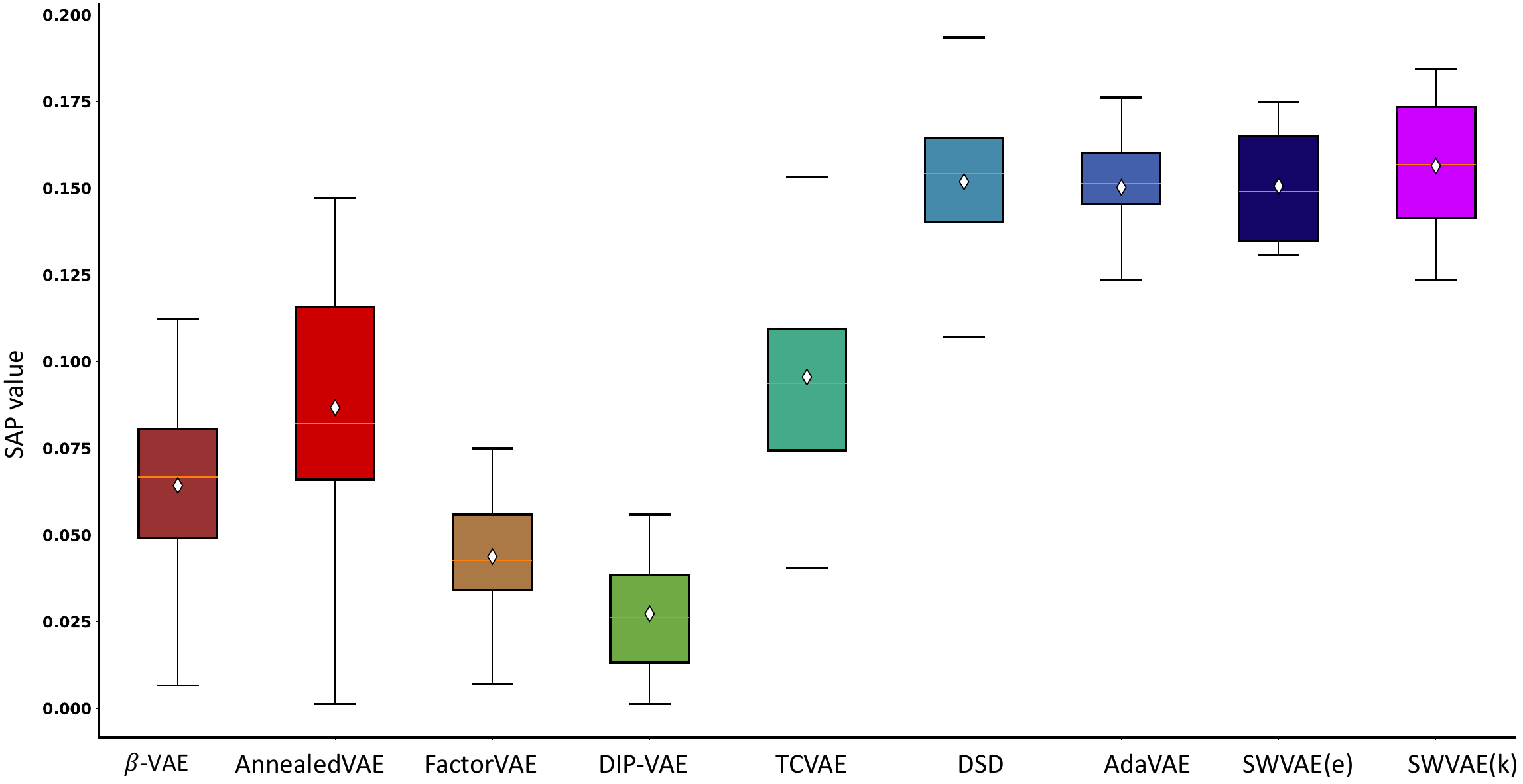} &
\includegraphics[width=0.3\textwidth,height=0.29\textwidth]{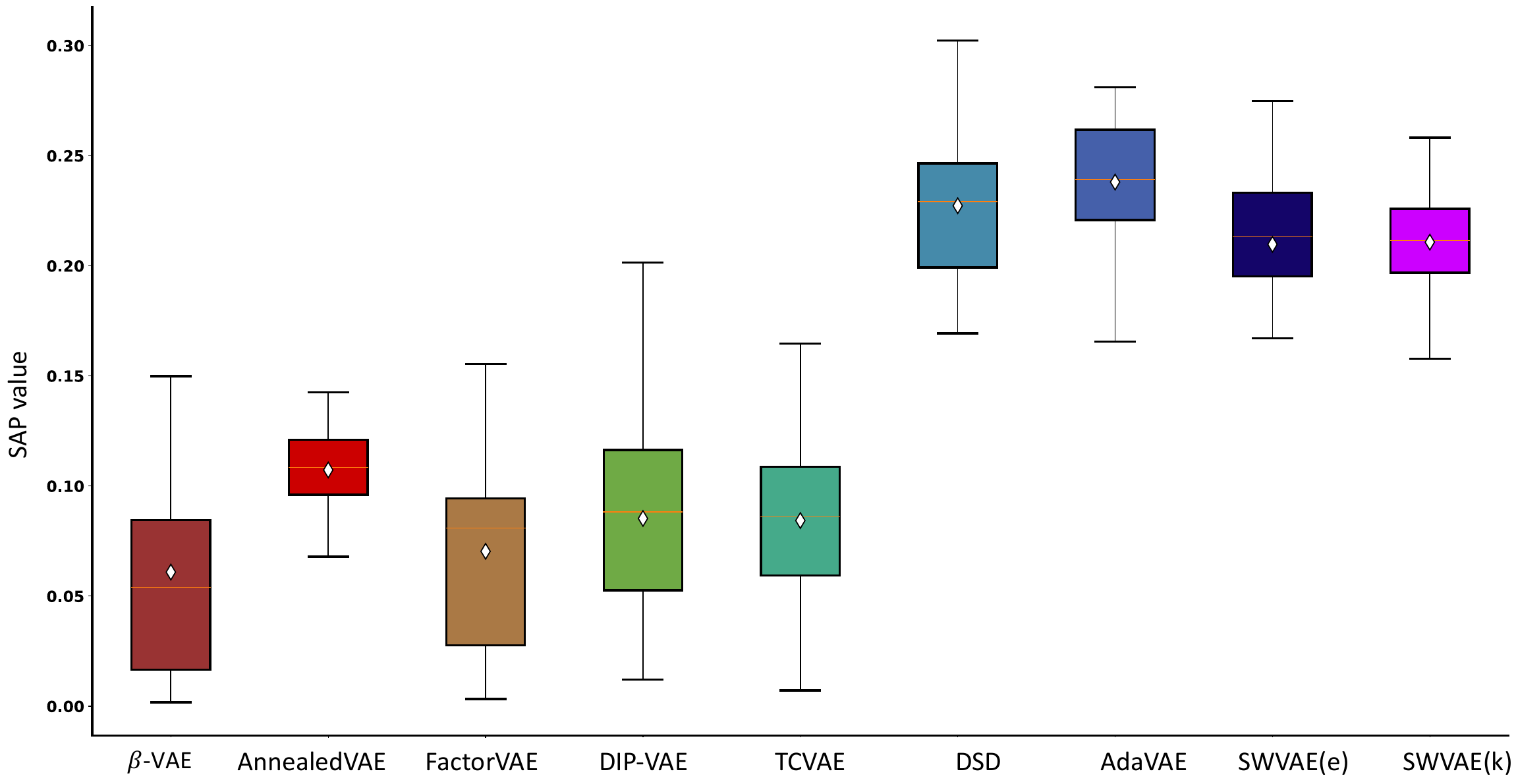} \\
\textbf{(a)}  & \textbf{(b)} & \textbf{(c)}  \\[6pt]
\end{tabular}
\begin{tabular}{cccc}
\includegraphics[width=0.3\textwidth,height=0.29\textwidth]{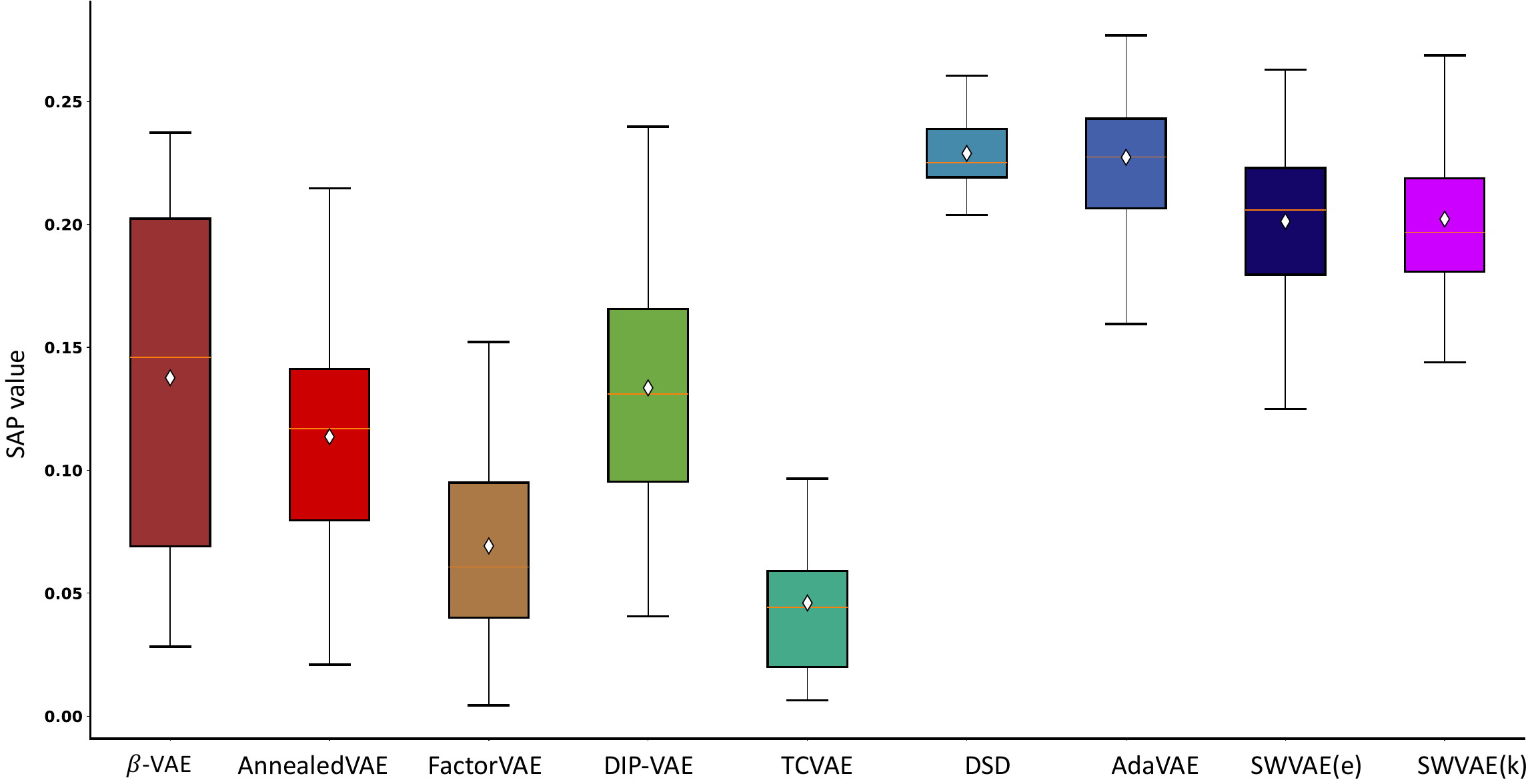} &
\includegraphics[width=0.3\textwidth,height=0.29\textwidth]{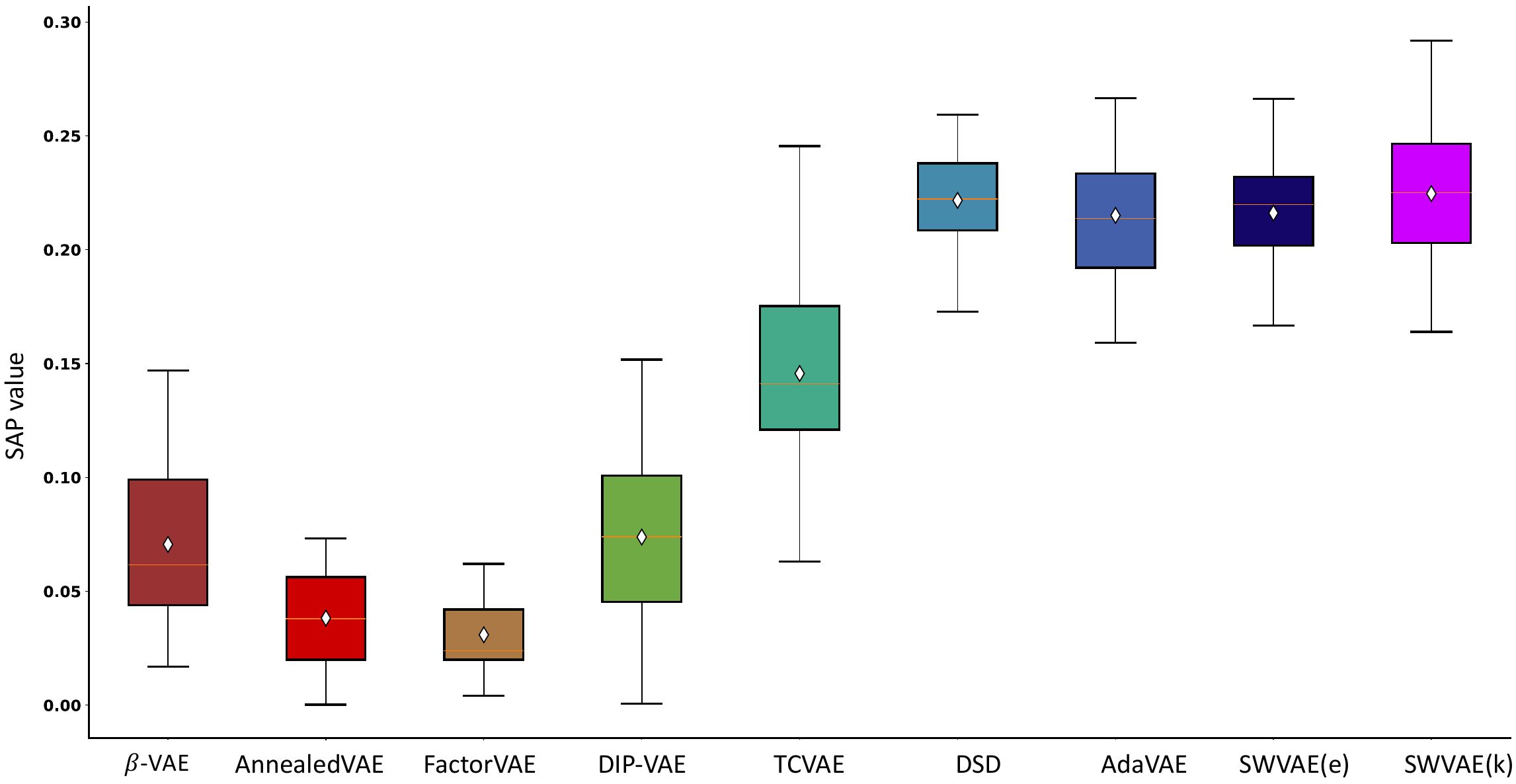} \\
\textbf{(d)}  & \textbf{(e)}  \\[6pt]
\end{tabular}
\centering
\caption{Box plot of SAP scores among models tested on different datasets. 
\textbf{(a)} dSprites 
\textbf{(b)} 3dSahpes
\textbf{(c)} MPI3D-toy
\textbf{(d)} MPI3D-realistic
\textbf{(e)} MPI3D-real
}
\label{fig:SAP}
\end{figure*}

\begin{figure*}[]

\centering
\begin{tabular}{cccc}
\includegraphics[width=0.3\textwidth,height=0.29\textwidth]{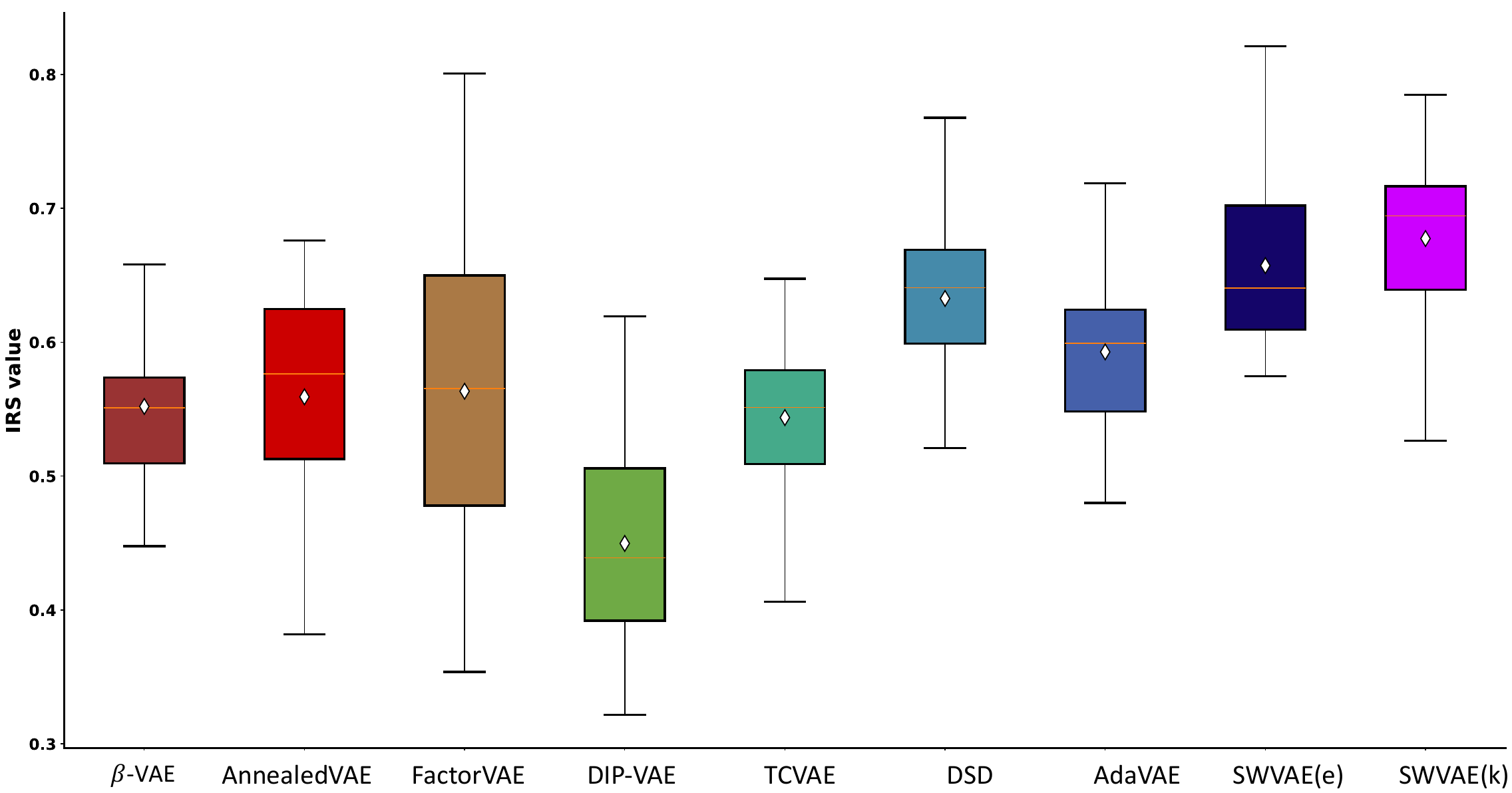} &
\includegraphics[width=0.3\textwidth,height=0.29\textwidth]{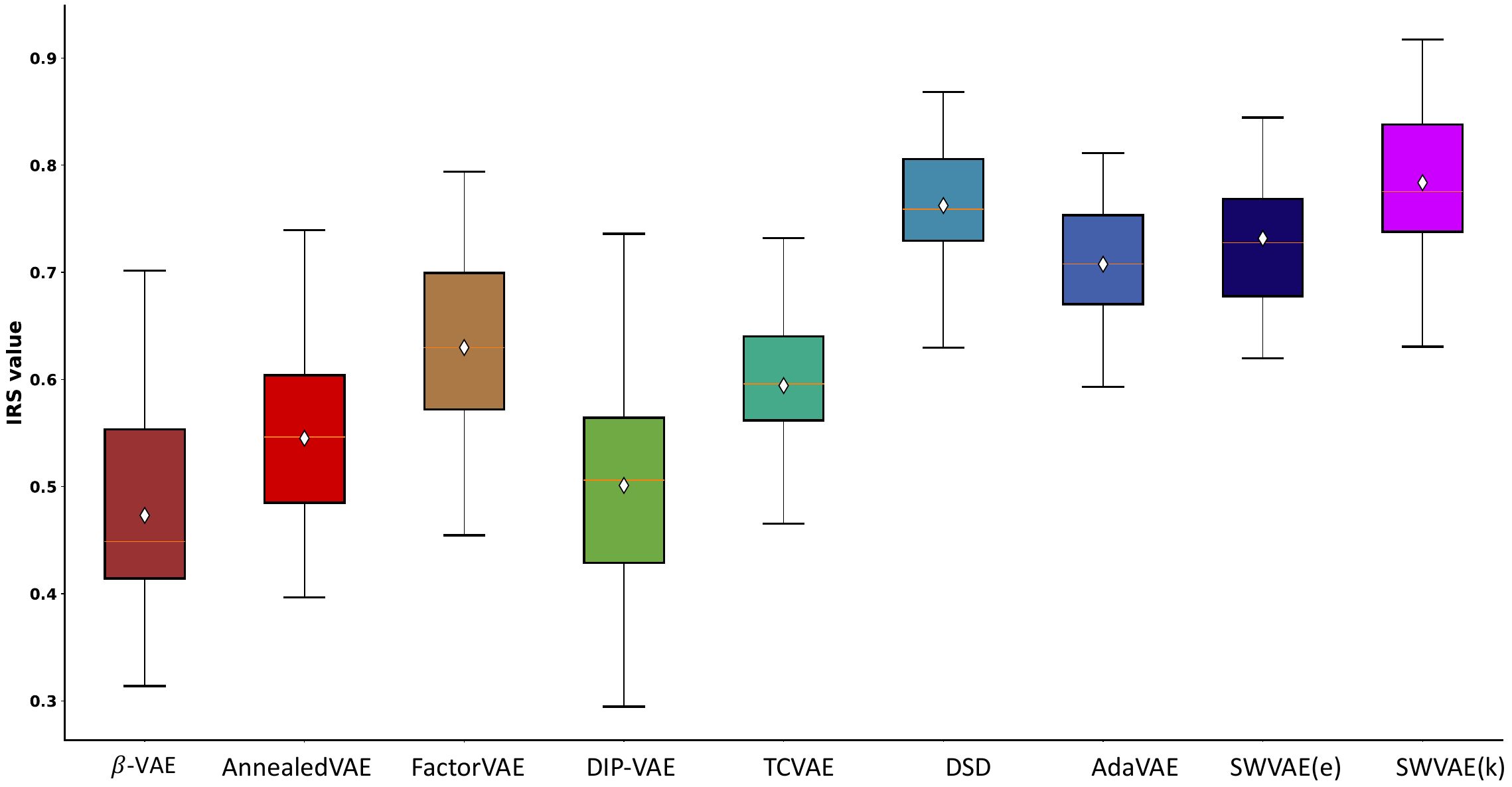} &
\includegraphics[width=0.3\textwidth,height=0.29\textwidth]{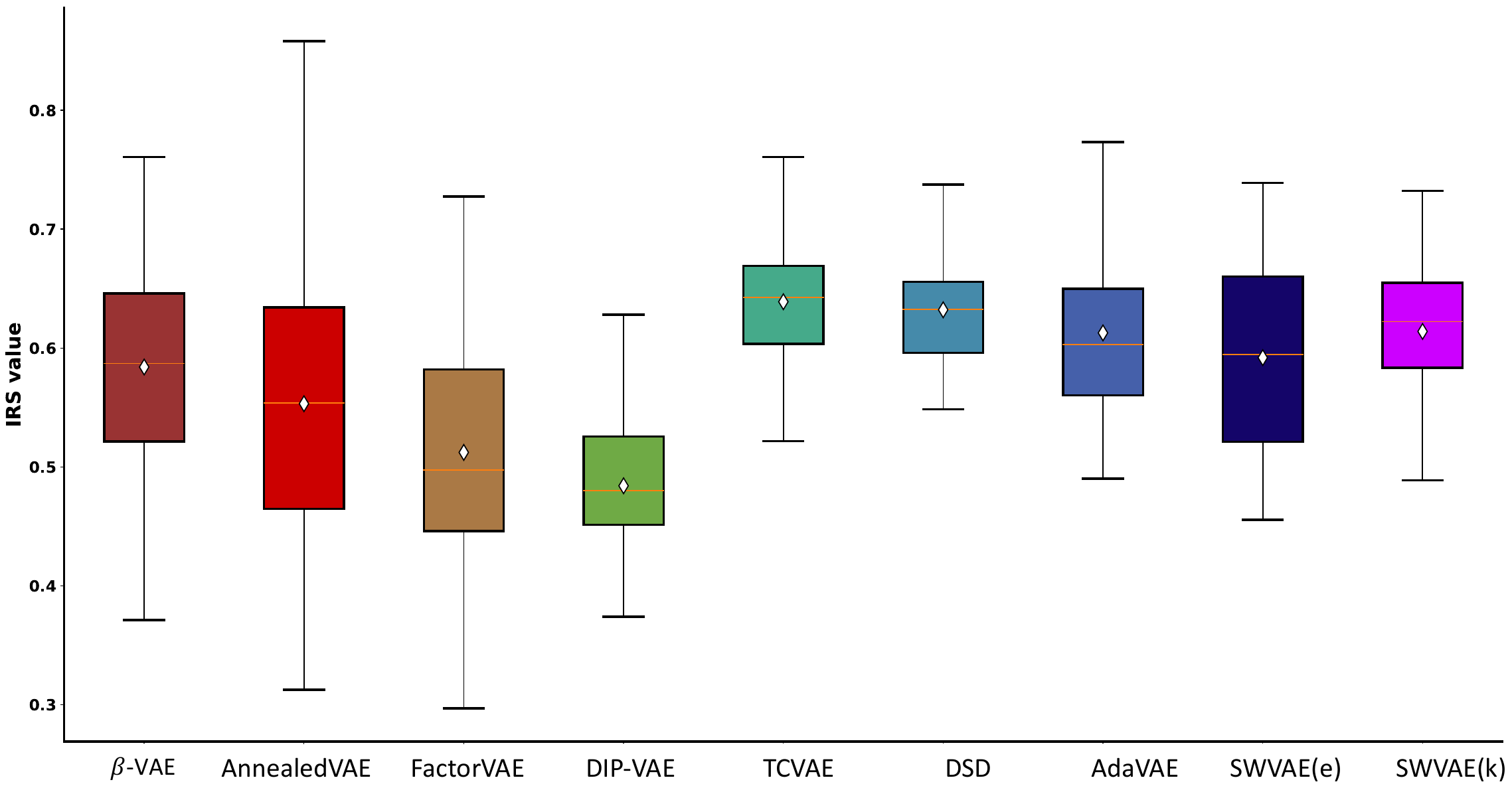} \\
\textbf{(a)}  & \textbf{(b)} & \textbf{(c)}  \\[6pt]
\end{tabular}
\begin{tabular}{cccc}
\includegraphics[width=0.3\textwidth,height=0.29\textwidth]{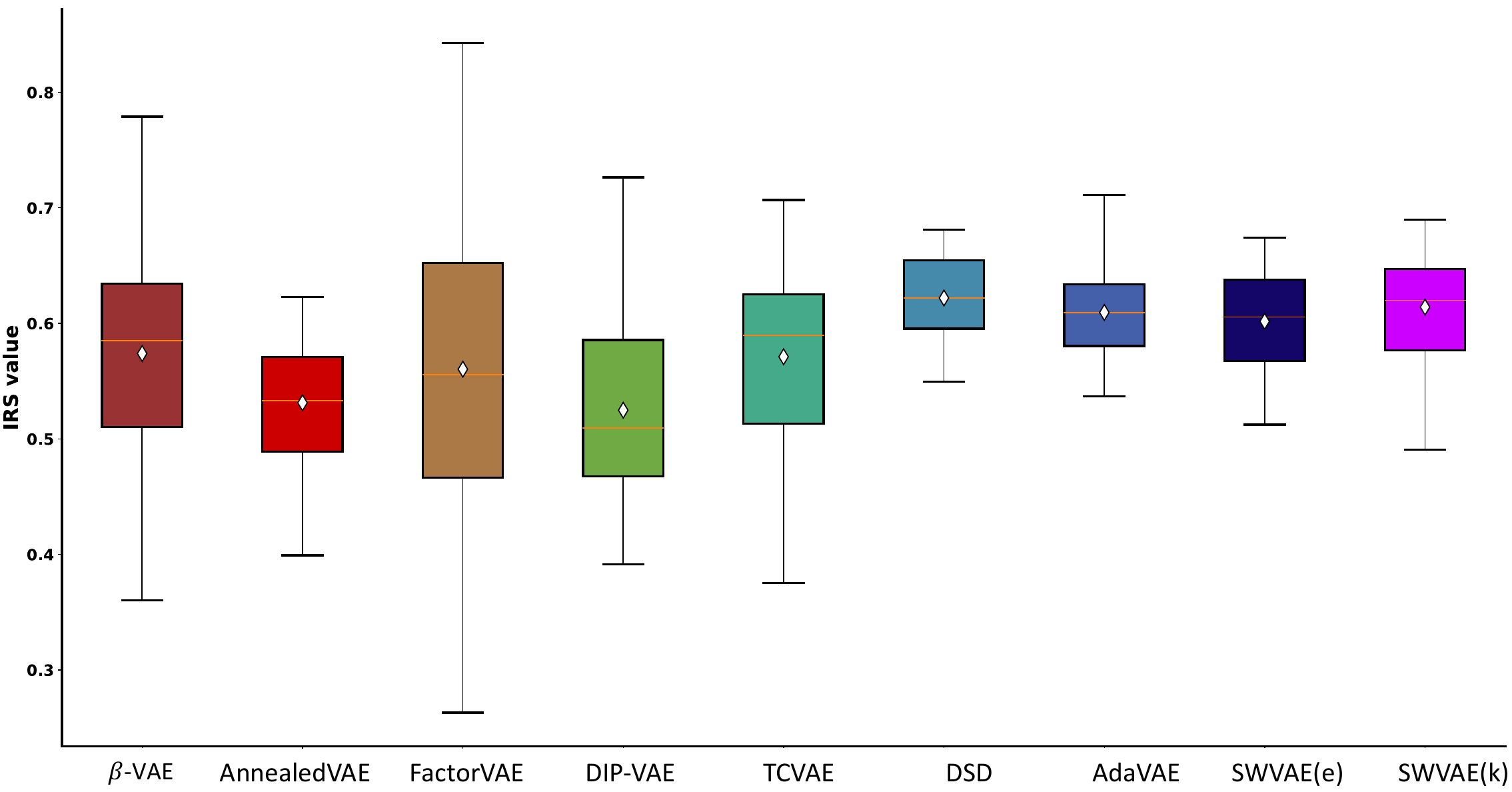} &
\includegraphics[width=0.3\textwidth,height=0.29\textwidth]{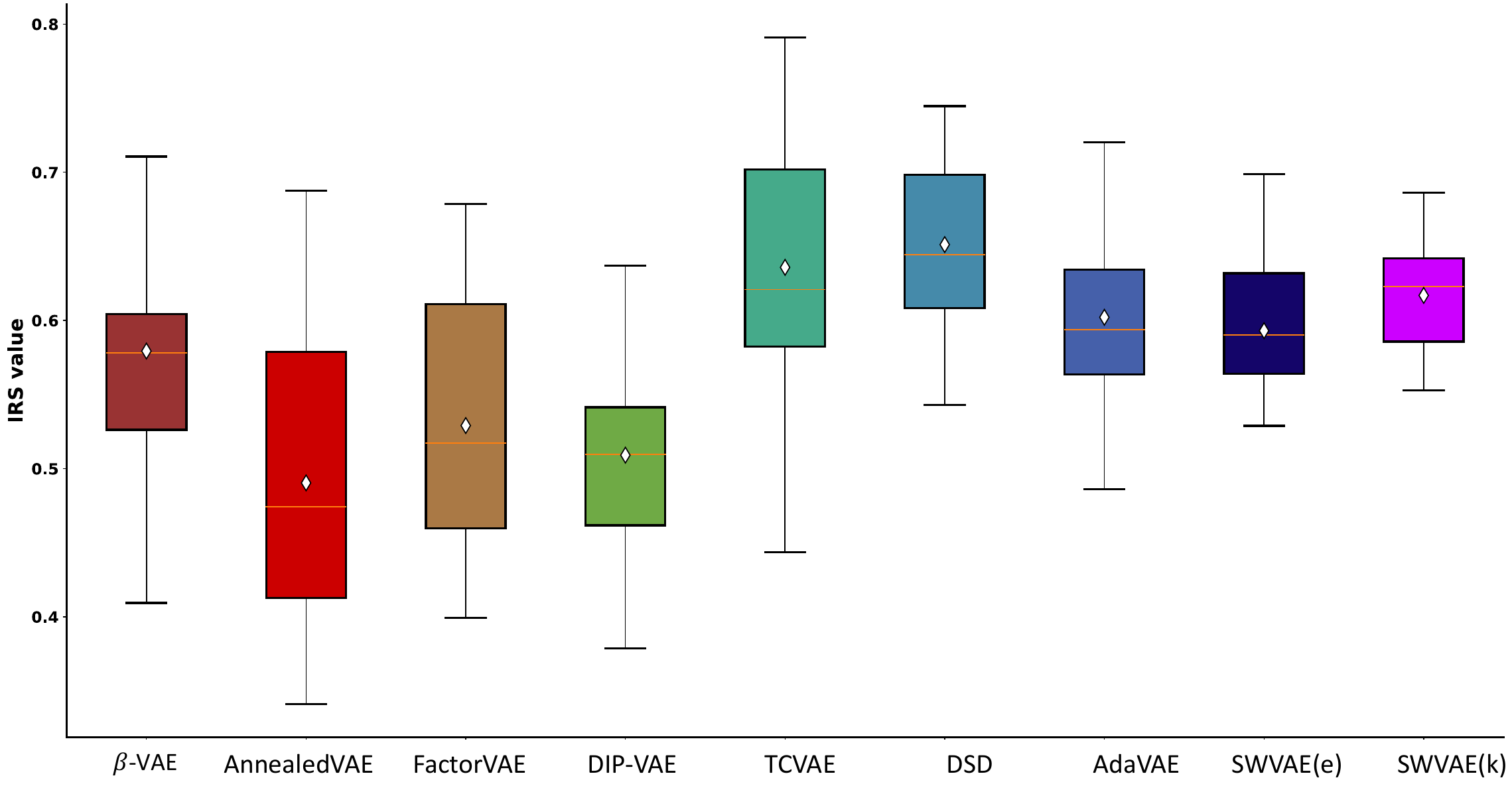} \\
\textbf{(d)}  & \textbf{(e)}  \\[6pt]
\end{tabular}
\centering
\caption{Box plot of IRS scores among models tested on different datasets. 
\textbf{(a)} dSprites 
\textbf{(b)} 3dSahpes
\textbf{(c)} MPI3D-toy
\textbf{(d)} MPI3D-realistic
\textbf{(e)} MPI3D-real
}
\label{fig:IRS}
\end{figure*}

\begin{figure*}[]

\centering
\begin{tabular}{cccc}
\includegraphics[width=0.3\textwidth,height=0.29\textwidth]{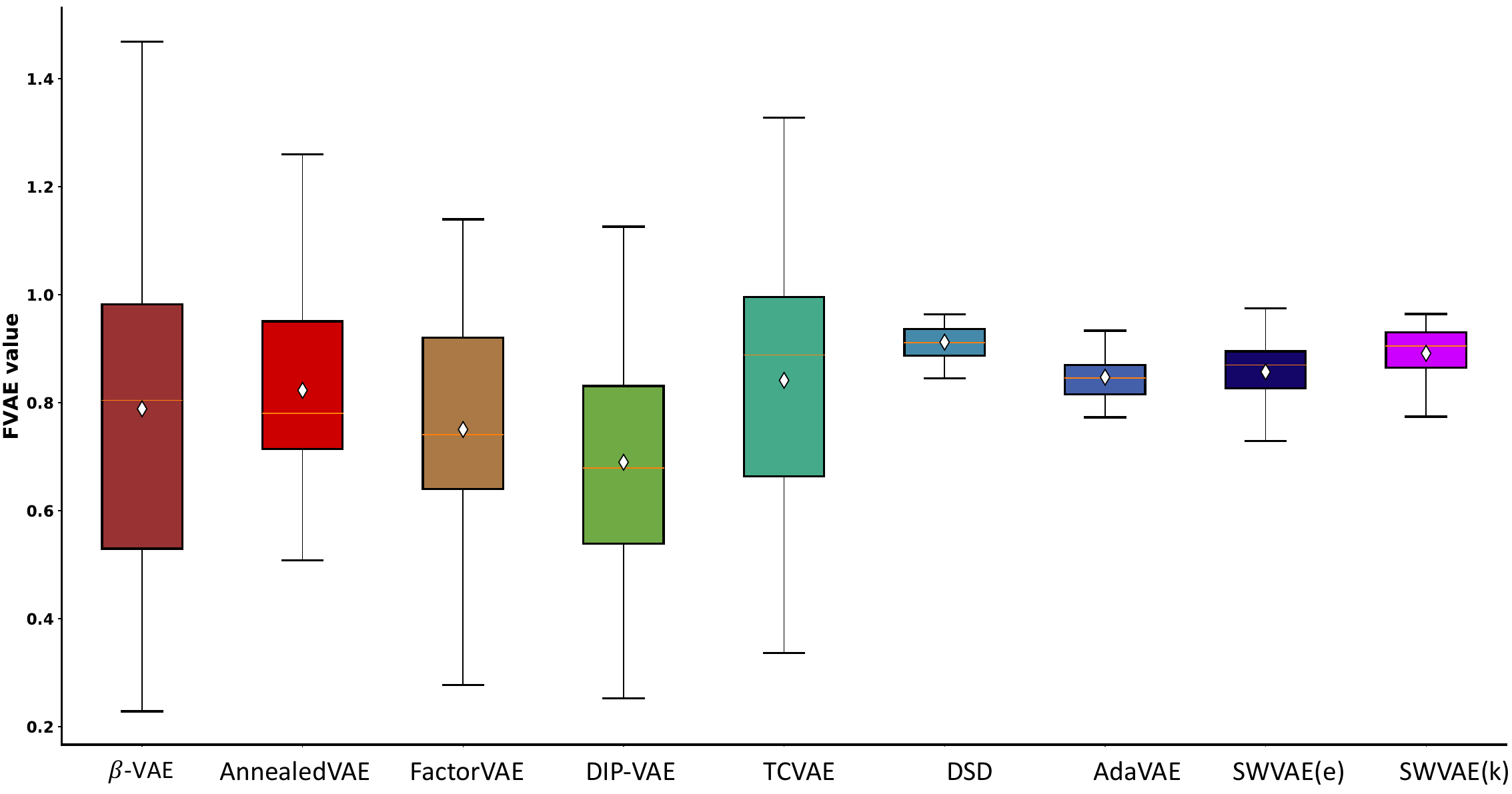} &
\includegraphics[width=0.3\textwidth,height=0.29\textwidth]{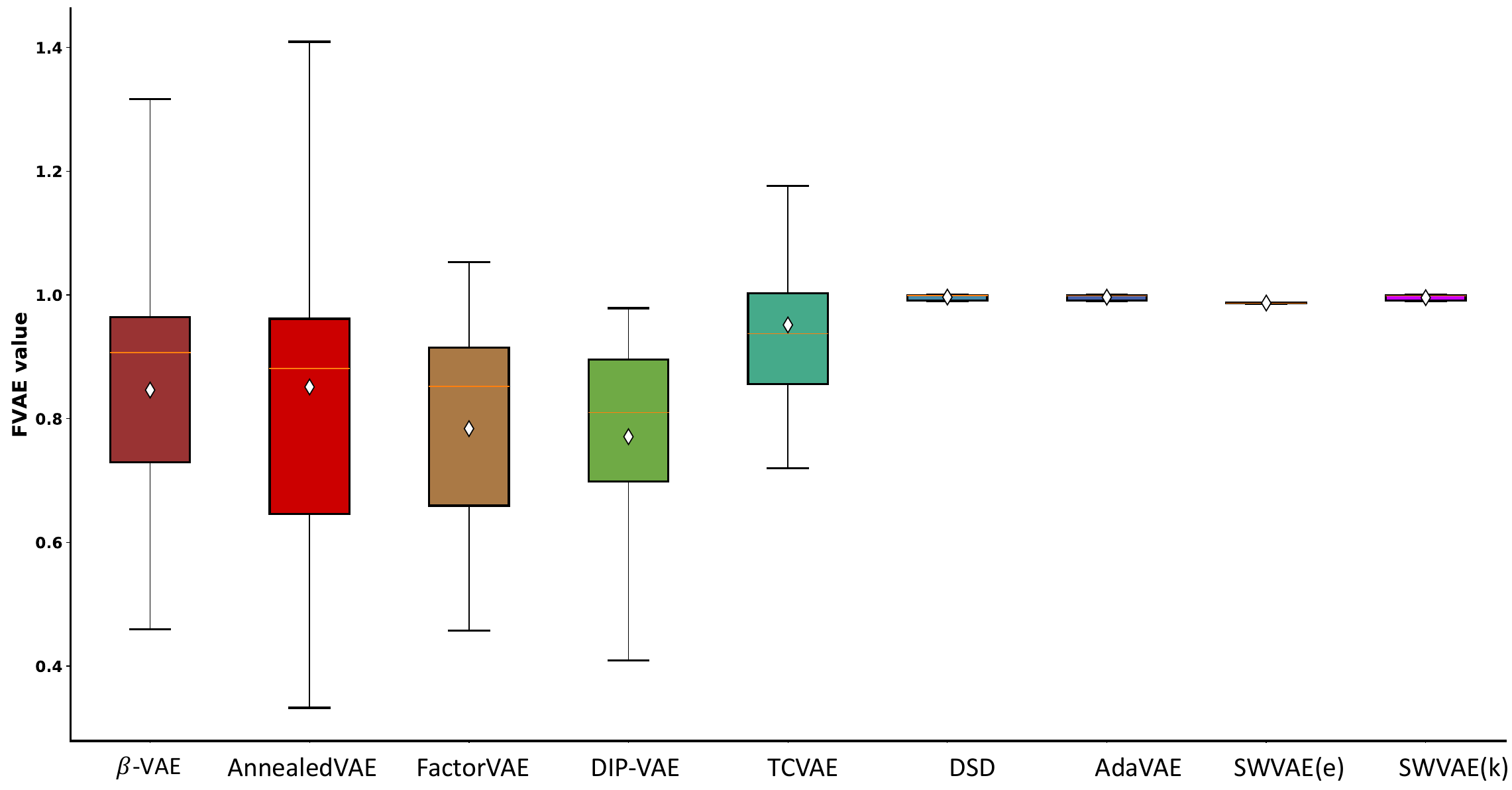} &
\includegraphics[width=0.3\textwidth,height=0.29\textwidth]{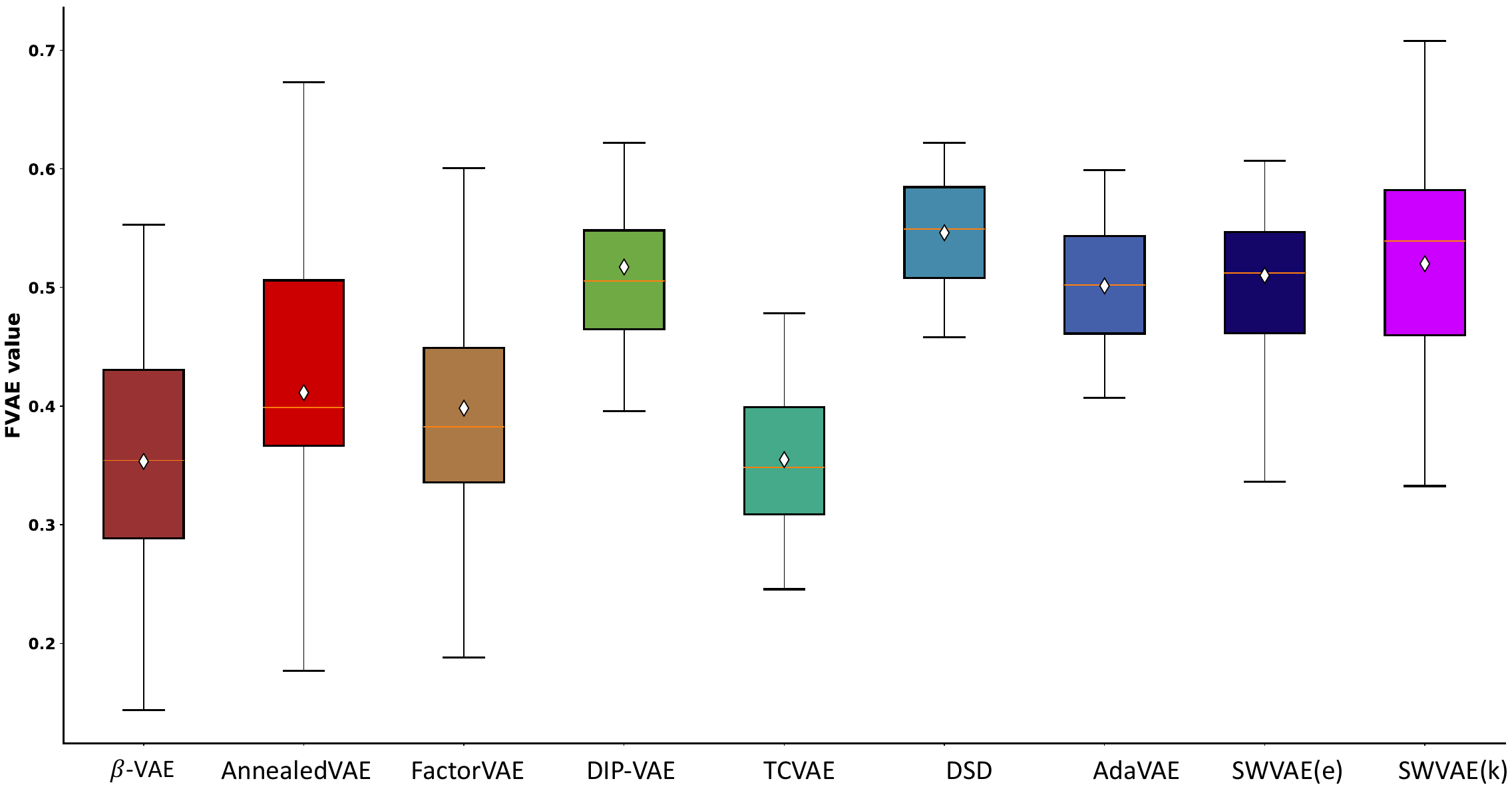} \\
\textbf{(a)}  & \textbf{(b)} & \textbf{(c)}  \\[6pt]
\end{tabular}
\begin{tabular}{cccc}
\includegraphics[width=0.3\textwidth,height=0.29\textwidth]{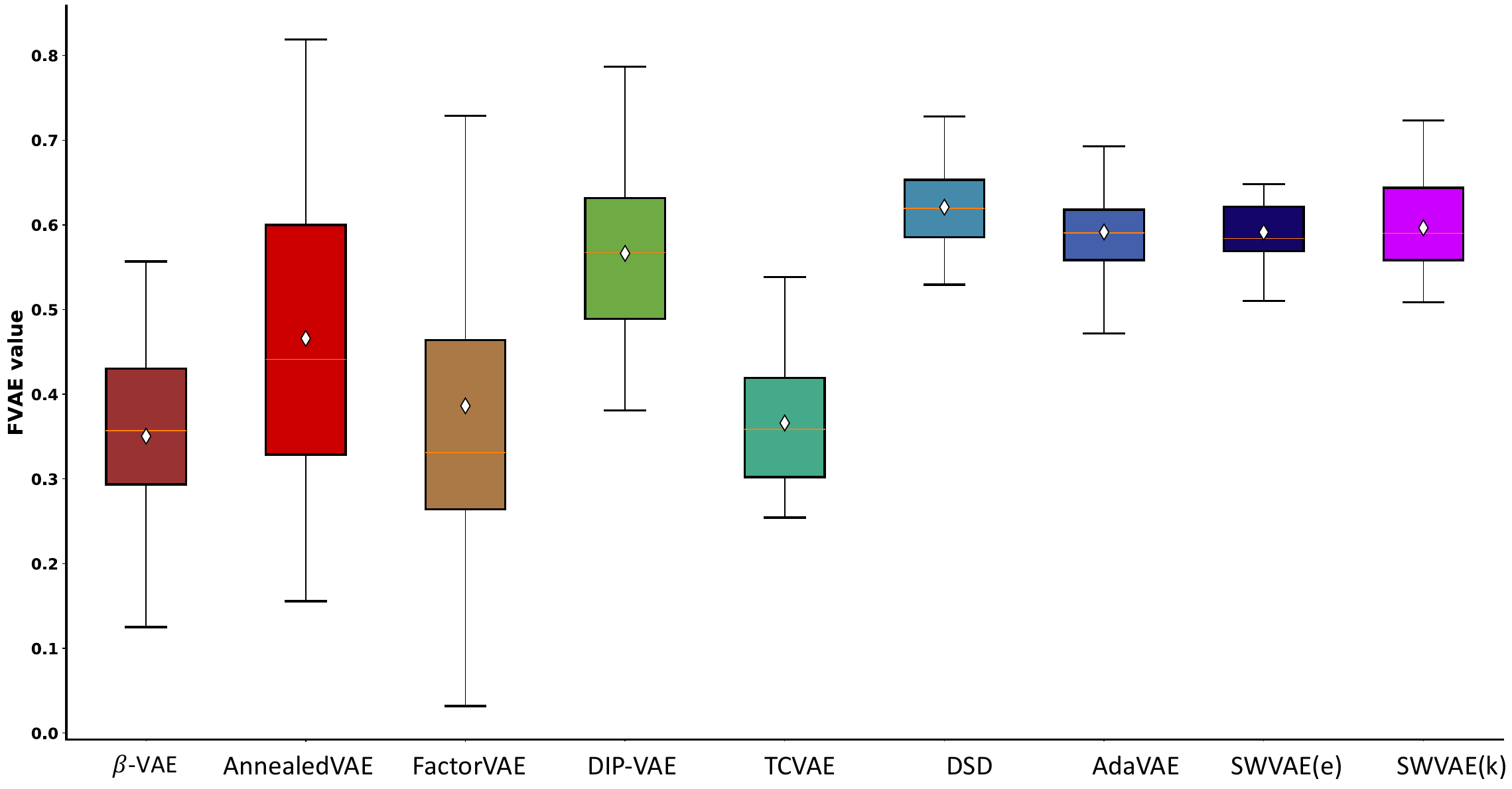} &
\includegraphics[width=0.3\textwidth,height=0.29\textwidth]{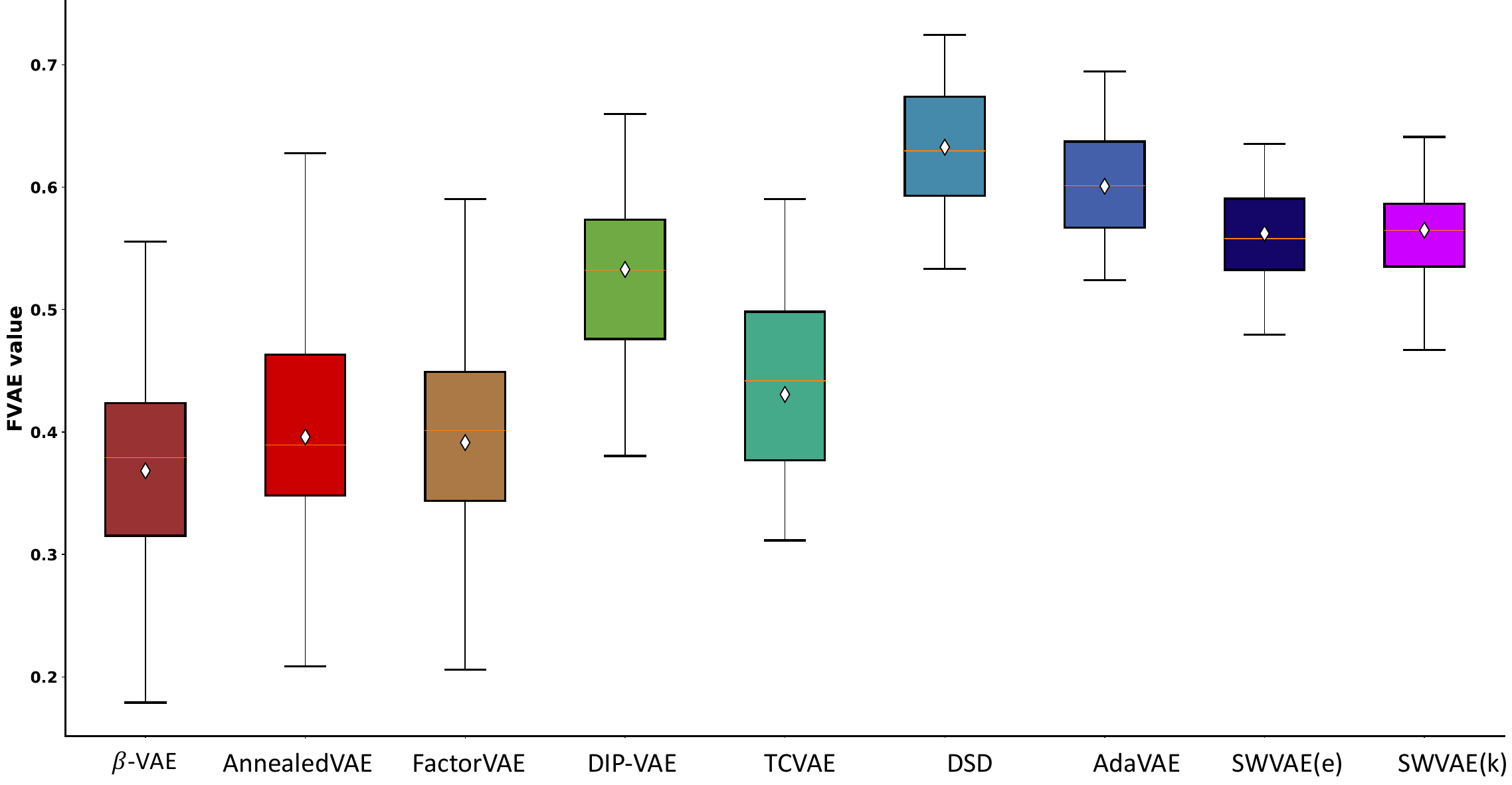} \\
\textbf{(d)}  & \textbf{(e)}   \\[6pt]
\end{tabular}
\centering
\caption{Box plot of FactorVAE scores among models tested on different datasets. 
\textbf{(a)} dSprites 
\textbf{(b)} 3dSahpes
\textbf{(c)} MPI3D-toy
\textbf{(d)} MPI3D-realistic
\textbf{(e)} MPI3D-real
}
\label{fig:FVAE}
\end{figure*}

\begin{figure*}[]

\centering
\begin{tabular}{cccc}
\includegraphics[width=0.3\textwidth,height=0.29\textwidth]{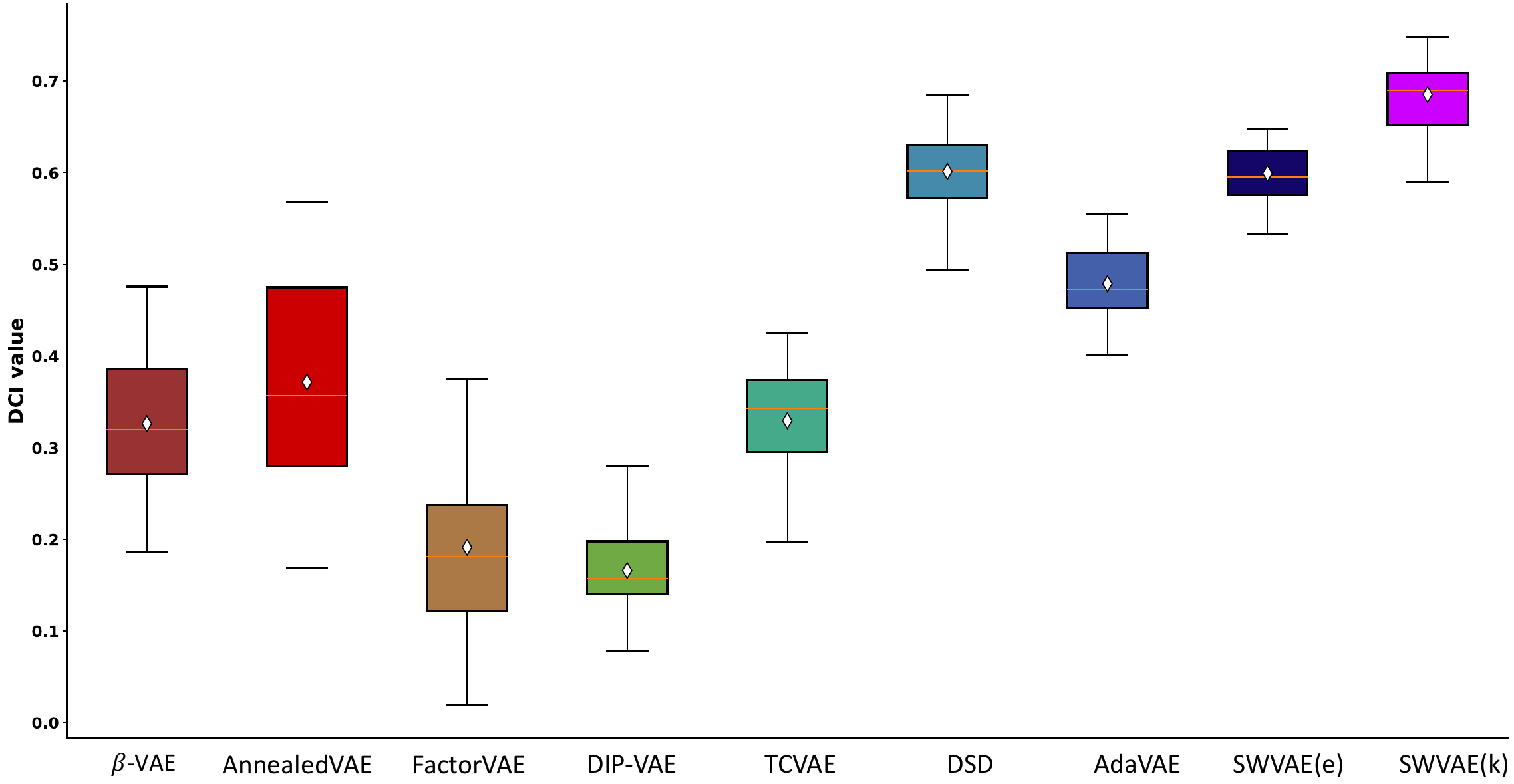} &
\includegraphics[width=0.3\textwidth,height=0.29\textwidth]{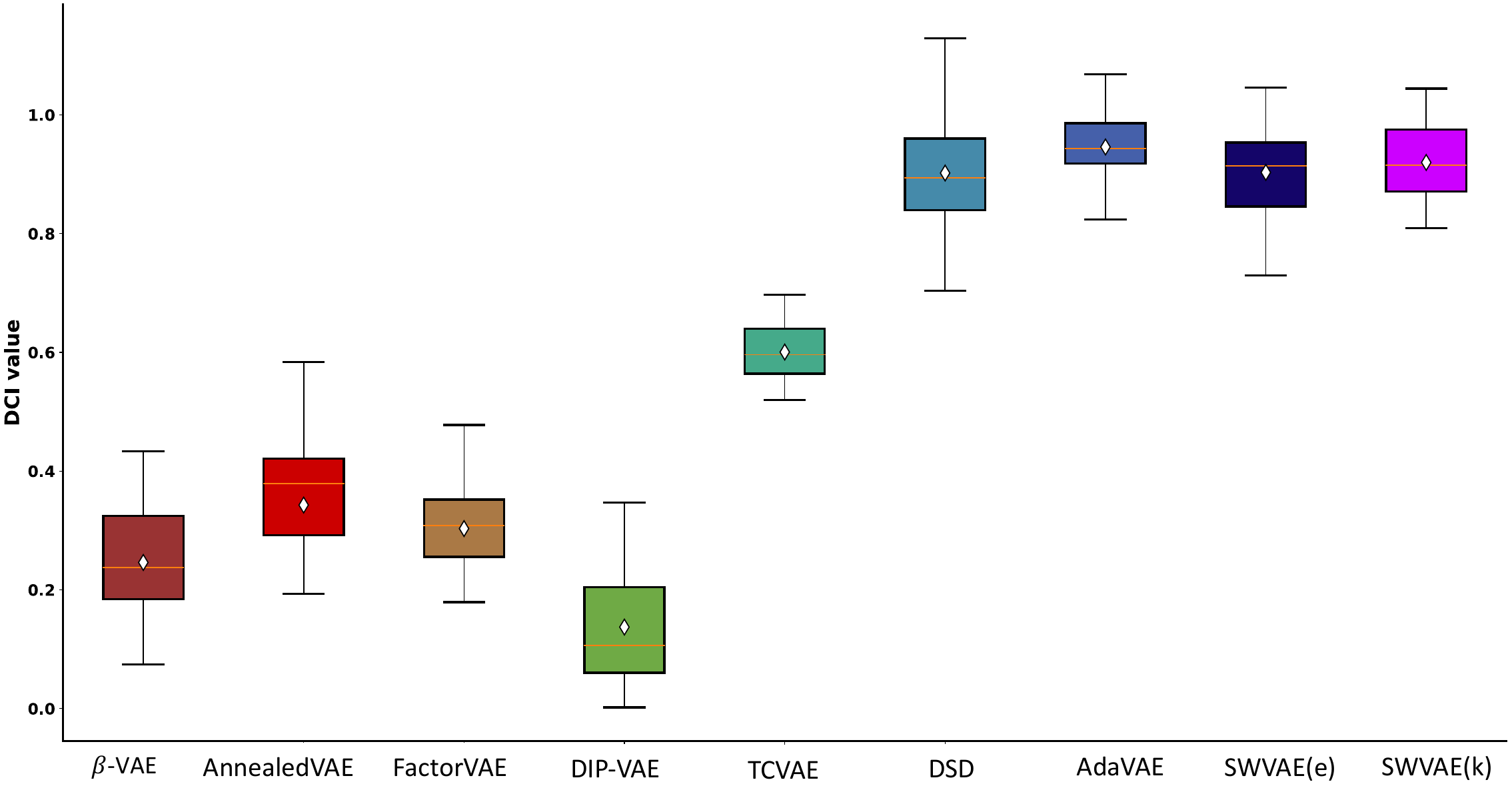} &
\includegraphics[width=0.3\textwidth,height=0.29\textwidth]{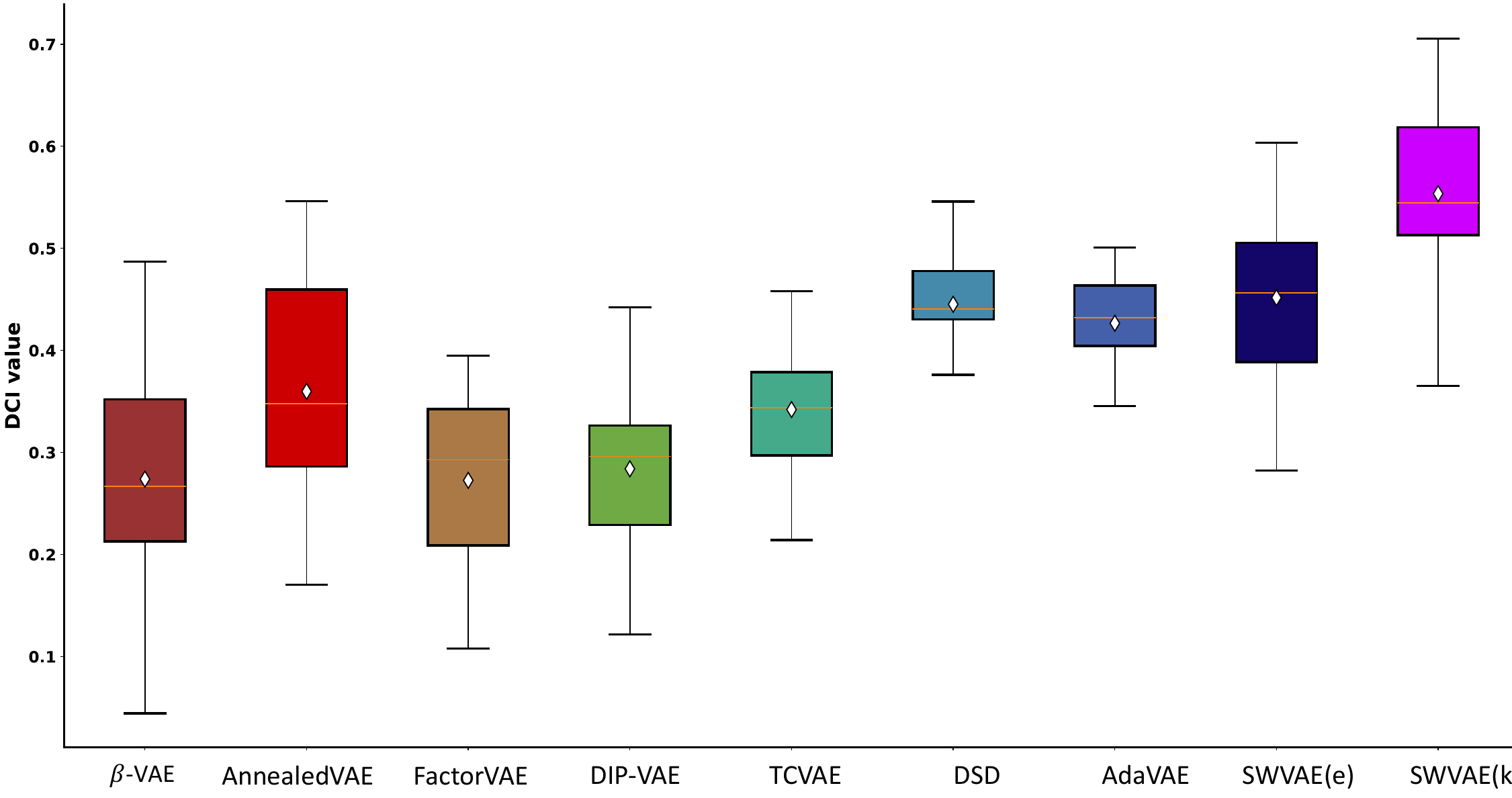} \\
\textbf{(a)}  & \textbf{(b)} & \textbf{(c)}  \\[6pt]
\end{tabular}
\begin{tabular}{cccc}
\includegraphics[width=0.3\textwidth,height=0.29\textwidth]{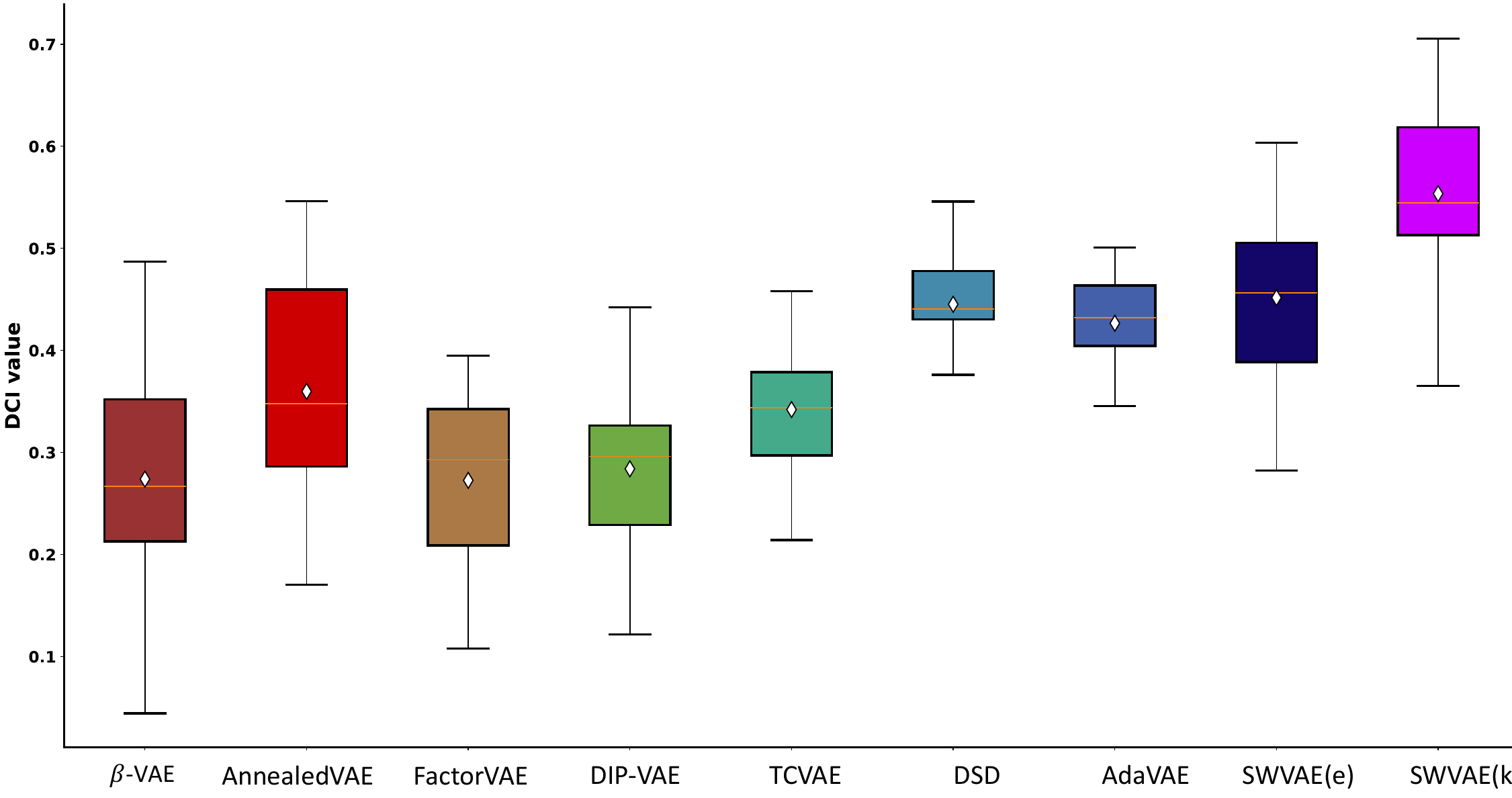} &
\includegraphics[width=0.3\textwidth,height=0.29\textwidth]{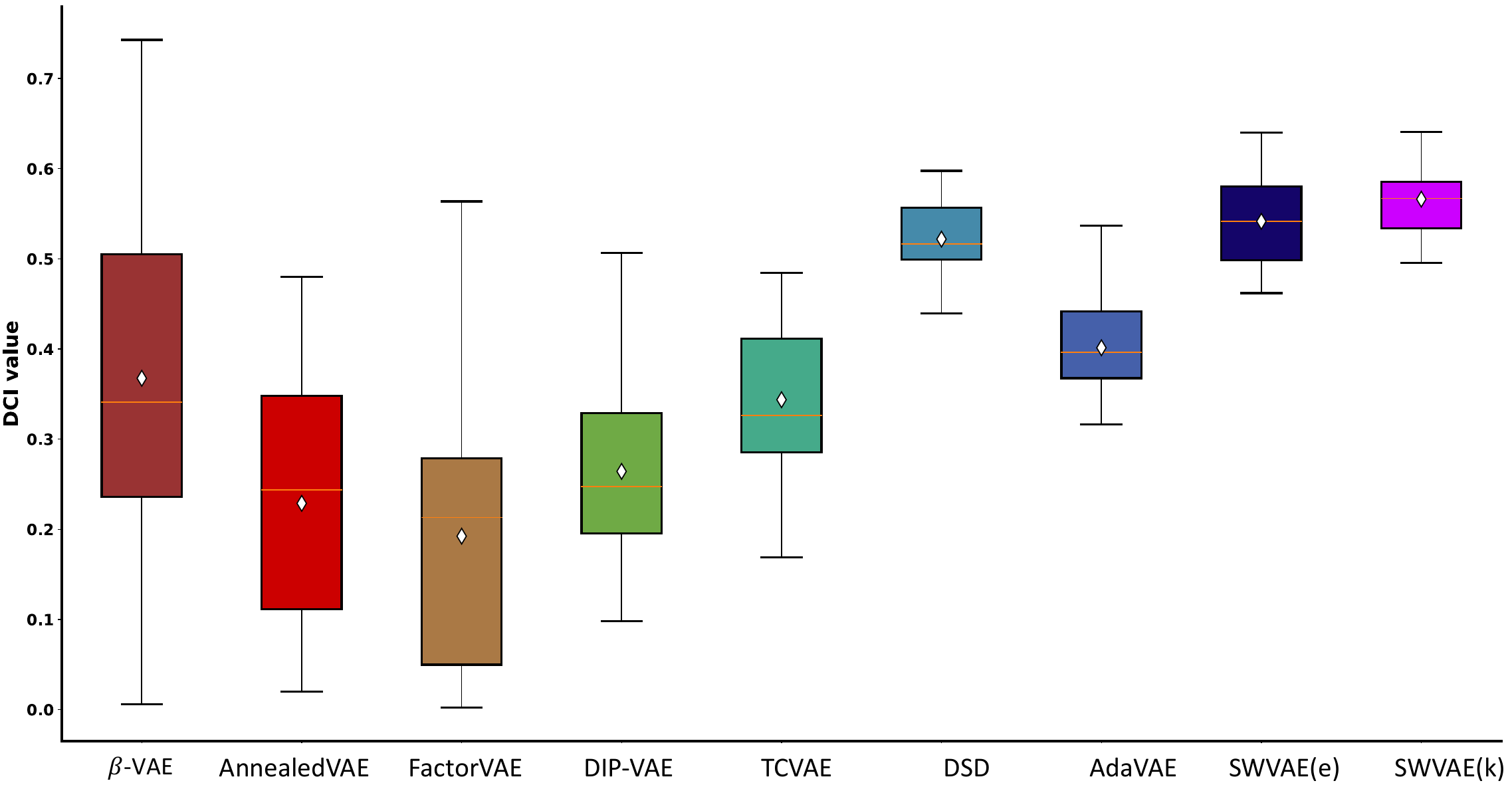} \\
\textbf{(d)}  & \textbf{(e)}   \\[6pt]
\end{tabular}
\centering
\caption{Box plot of DCI scores among models tested on different datasets. 
\textbf{(a)} dSprites 
\textbf{(b)} 3dSahpes
\textbf{(c)} MPI3D-toy
\textbf{(d)} MPI3D-realistic
\textbf{(e)} MPI3D-real
}
\label{fig:DCI}
\end{figure*}

We show the box plot of MIG scores on all datasets  in \cref{fig:MIG}, SAP scores on all datasets \cref{fig:SAP}, IRS scores on all datasets in \cref{fig:IRS}, FactorVAE scores on all datasets in \cref{fig:FVAE}, DCI scores on all datasets \cref{fig:DCI}.

\subsection{Traversal visualization}
\begin{figure*}[]

\centering
\begin{tabular}{cccc}
\includegraphics[width=0.3\textwidth,height=0.29\textwidth]{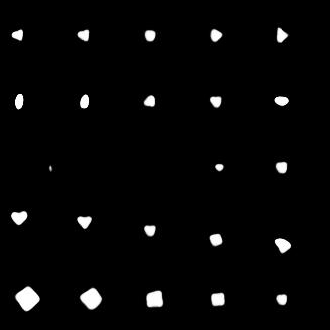} &
\includegraphics[width=0.3\textwidth,height=0.29\textwidth]{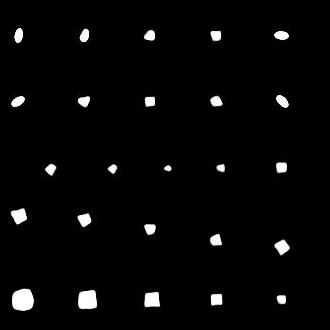} &
\includegraphics[width=0.3\textwidth,height=0.29\textwidth]{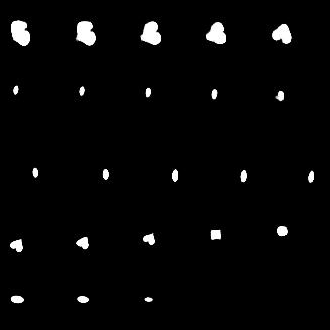} \\
\textbf{(a)}  & \textbf{(b)} & \textbf{(c)}  \\[6pt]
\end{tabular}
\begin{tabular}{cccc}
\includegraphics[width=0.3\textwidth,height=0.29\textwidth]{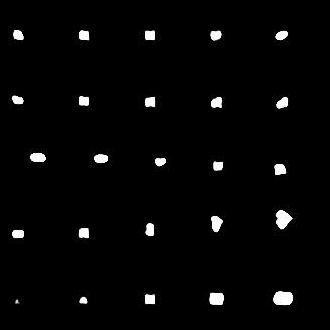} &
\includegraphics[width=0.3\textwidth,height=0.29\textwidth]{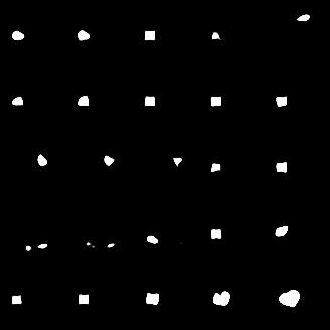} &
\includegraphics[width=0.3\textwidth,height=0.29\textwidth]{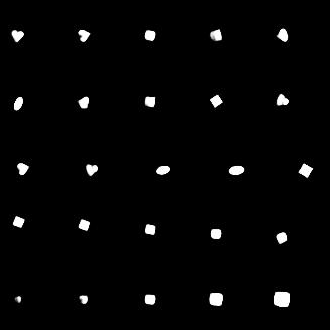} \\
\textbf{(d)}  & \textbf{(e)} & \textbf{(f)}  \\[6pt]
\end{tabular}
\begin{tabular}{cccc}
\includegraphics[width=0.3\textwidth,height=0.29\textwidth]{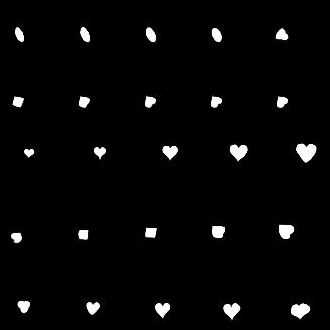} &
\includegraphics[width=0.3\textwidth,height=0.29\textwidth]{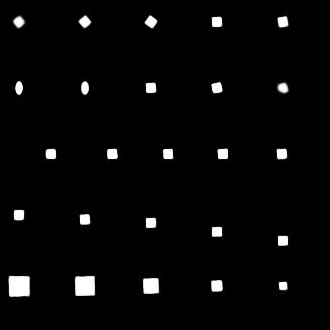} \\
\textbf{(g)}  & \textbf{(h)}  \\[6pt]
\end{tabular}
\centering
\caption{dSprites dataset traverse visualization. Each row of sub-figure represents: rotation, shape, x-position, y-position and size.
\textbf{(a)} AnnealedVAE 
\textbf{(b)} $\beta$-VAE
\textbf{(c)} FactorVAE
\textbf{(d)} DIPVAEI
\textbf{(e)} DIPVAEII
\textbf{(f)} $\beta$-TCVAE
\textbf{(g)} $\beta$-FactorTCVAE
\textbf{(h)} \textbf{SW-VAE}
}
\label{fig:dsprites_traverse}
\end{figure*}

\begin{figure*}[!htb]
\centering
\begin{tabular}{cccc}
\includegraphics[width=0.3\textwidth]{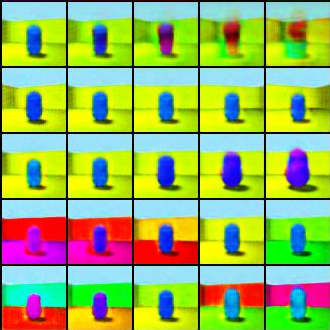} &
\includegraphics[width=0.3\textwidth]{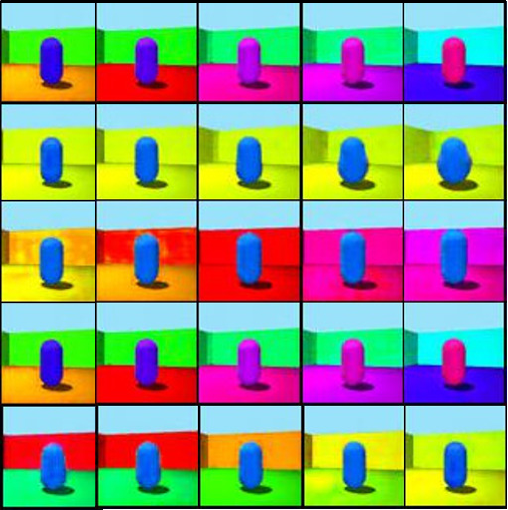} &
\includegraphics[width=0.3\textwidth]{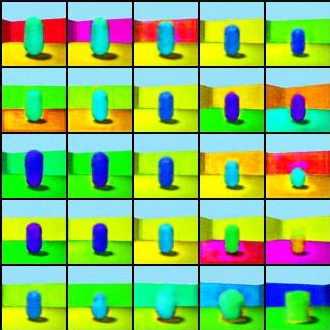} \\
\textbf{(a)}  & \textbf{(b)} & \textbf{(c)}  \\[6pt]
\end{tabular}
\begin{tabular}{cccc}
\includegraphics[width=0.3\textwidth]{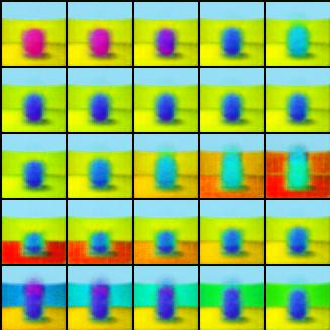} &
\includegraphics[width=0.3\textwidth]{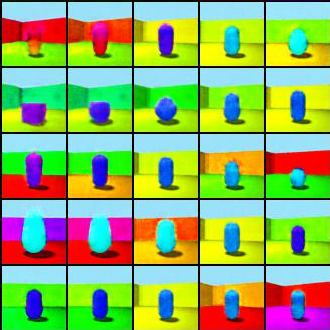} &
\includegraphics[width=0.3\textwidth]{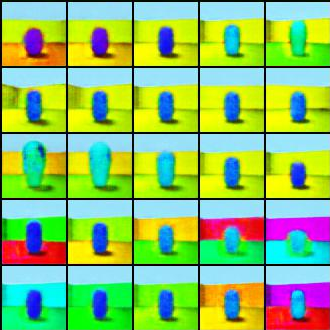} \\
\textbf{(d)}  & \textbf{(e)} & \textbf{(f)}  \\[6pt]
\end{tabular}
\begin{tabular}{cccc}
\includegraphics[width=0.3\textwidth]{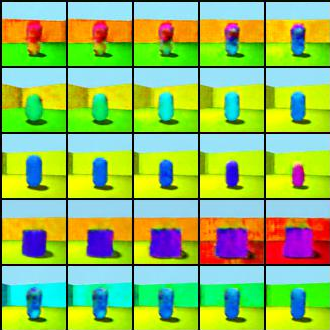} &
\includegraphics[width=0.3\textwidth]{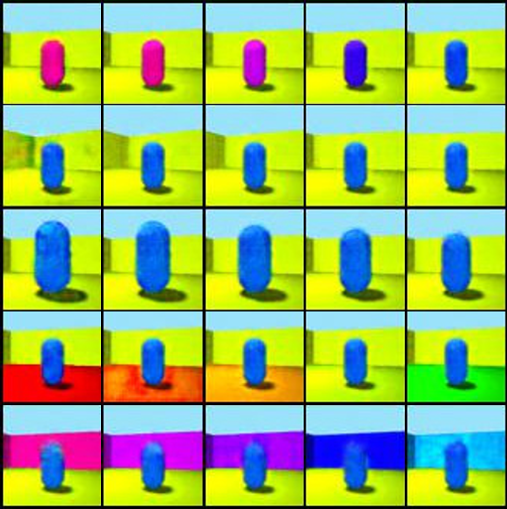} \\
\textbf{(g)}  & \textbf{(h)}  \\[6pt]
\end{tabular}
\centering
\caption{shapes3d dataset traverse visualization. Each row of sub-figure represents: object-color, orientation, size, floor color  and wall-color.
\textbf{(a)} AnnealedVAE 
\textbf{(b)} $\beta$-VAE
\textbf{(c)} FactorVAE
\textbf{(d)} DIPVAEI
\textbf{(e)} DIPVAEII
\textbf{(f)} $\beta$-TCVAE
\textbf{(g)} $\beta$-FactorTCVAE
\textbf{(h)} \textbf{SW-VAE}
}

\label{fig:shapes3d_traverse}
\end{figure*}

\begin{figure*}[!htb]
\centering
\begin{tabular}{cccc}
\includegraphics[width=0.3\textwidth]{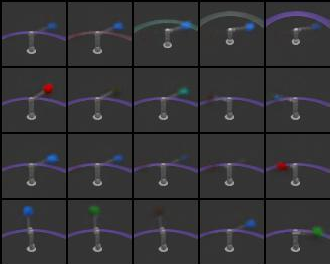} &
\includegraphics[width=0.3\textwidth]{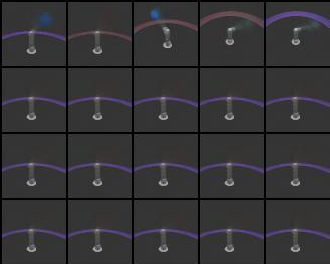} &
\includegraphics[width=0.3\textwidth]{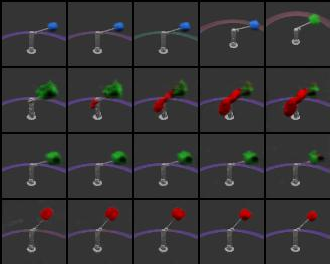} \\
\textbf{(a)}  & \textbf{(b)} & \textbf{(c)}  \\[6pt]
\end{tabular}
\begin{tabular}{cccc}
\includegraphics[width=0.3\textwidth]{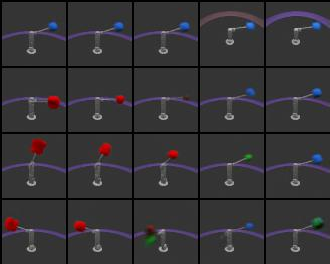} &
\includegraphics[width=0.3\textwidth]{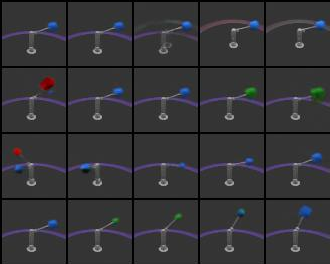} &
\includegraphics[width=0.3\textwidth]{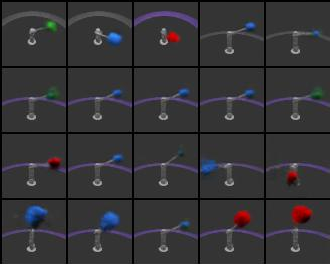} \\
\textbf{(d)}  & \textbf{(e)} & \textbf{(f)}  \\[6pt]
\end{tabular}
\begin{tabular}{cccc}
\includegraphics[width=0.3\textwidth]{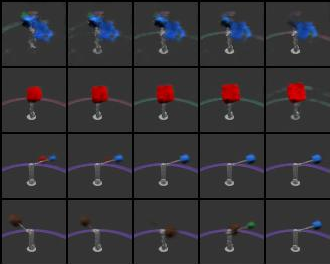} &
\includegraphics[width=0.3\textwidth]{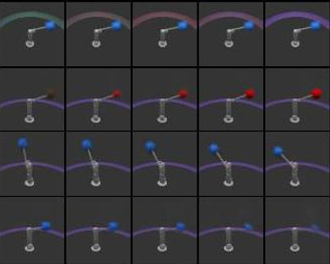} \\
\textbf{(g)}  & \textbf{(h)}  \\[6pt]
\end{tabular}
\centering
\caption{mpi3d-toy dataset traverse visualization. Each row of sub-figure represents: background-color, object-color, x-rotation, y-rotation.
\textbf{(a)} AnnealedVAE 
\textbf{(b)} $\beta$-VAE
\textbf{(c)} FactorVAE
\textbf{(d)} DIPVAEI
\textbf{(e)} DIPVAEII
\textbf{(f)} $\beta$-TCVAE
\textbf{(g)} $\beta$-FactorTCVAE
\textbf{(h)} \textbf{SW-VAE}
}

\label{fig:mpi3d-toy_traverse}
\end{figure*}

\begin{figure*}[!htb]
\centering
\begin{tabular}{cccc}
\includegraphics[width=0.3\textwidth]{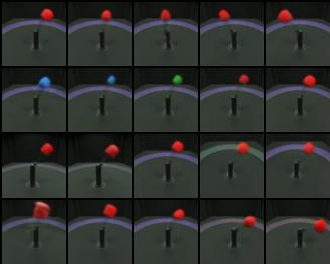} &
\includegraphics[width=0.3\textwidth]{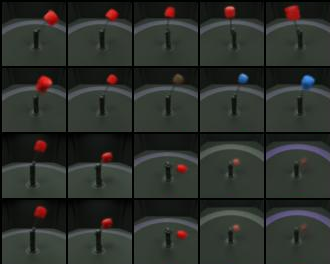} &
\includegraphics[width=0.3\textwidth]{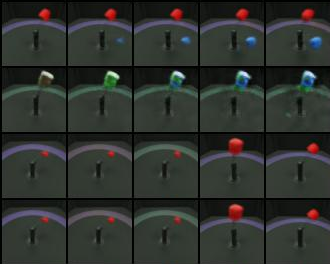} \\
\textbf{(a)}  & \textbf{(b)} & \textbf{(c)}  \\[6pt]
\end{tabular}
\begin{tabular}{cccc}
\includegraphics[width=0.3\textwidth]{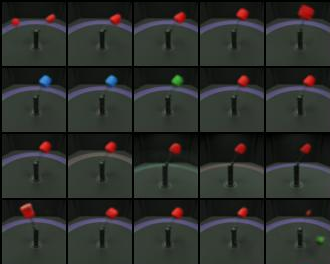} &
\includegraphics[width=0.3\textwidth]{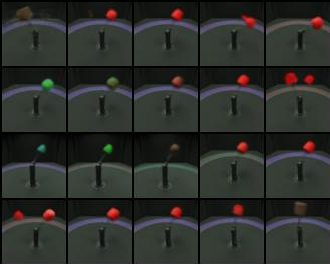} &
\includegraphics[width=0.3\textwidth]{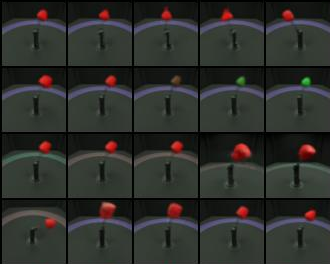} \\
\textbf{(d)}  & \textbf{(e)} & \textbf{(f)}  \\[6pt]
\end{tabular}
\begin{tabular}{cccc}
\includegraphics[width=0.3\textwidth]{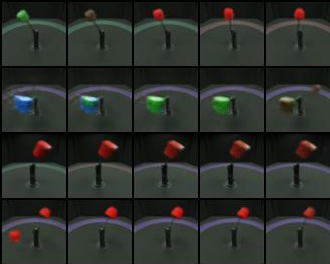} &
\includegraphics[width=0.3\textwidth]{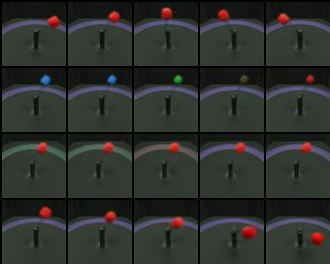} \\
\textbf{(g)}  & \textbf{(h)}  \\[6pt]
\end{tabular}
\centering
\caption{mpi3d-realistic dataset traverse visualization. Each row of sub-figure represents: x-rotation, object-color, background-color, y-rotation.
\textbf{(a)} AnnealedVAE 
\textbf{(b)} $\beta$-VAE
\textbf{(c)} FactorVAE
\textbf{(d)} DIPVAEI
\textbf{(e)} DIPVAEII
\textbf{(f)} $\beta$-TCVAE
\textbf{(g)} $\beta$-FactorTCVAE
\textbf{(h)} \textbf{SW-VAE}
}

\label{fig:mpi3d-realistic_traverse}
\end{figure*}

\begin{figure*}[!htb]
\centering
\begin{tabular}{cccc}
\includegraphics[width=0.3\textwidth]{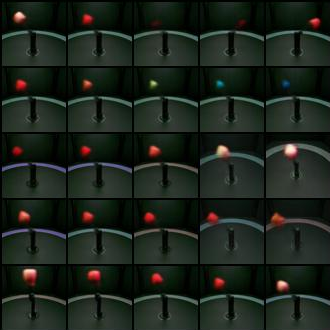} &
\includegraphics[width=0.3\textwidth]{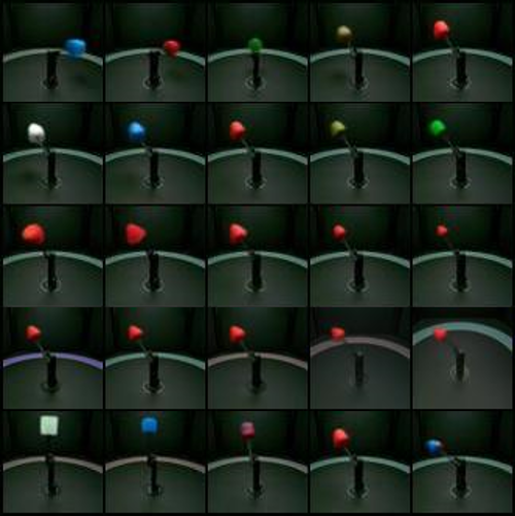} &
\includegraphics[width=0.3\textwidth]{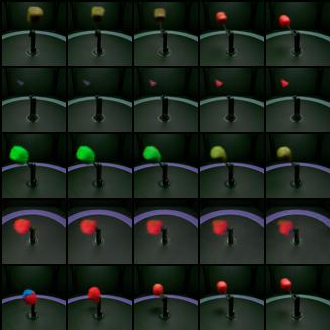} \\
\textbf{(a)}  & \textbf{(b)} & \textbf{(c)}  \\[6pt]
\end{tabular}
\centering
\begin{tabular}{cccc}
\includegraphics[width=0.3\textwidth]{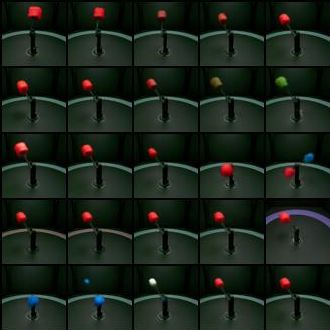} &
\includegraphics[width=0.3\textwidth]{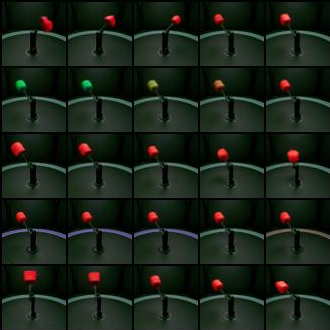} &
\includegraphics[width=0.3\textwidth]{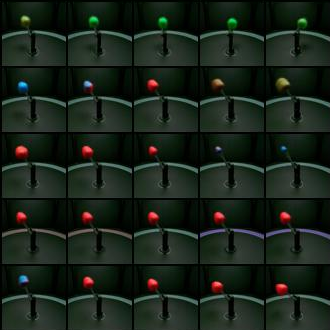} \\
\textbf{(d)}  & \textbf{(e)} & \textbf{(f)}  \\[6pt]
\centering
\end{tabular}
\begin{tabular}{cccc}
\includegraphics[width=0.3\textwidth]{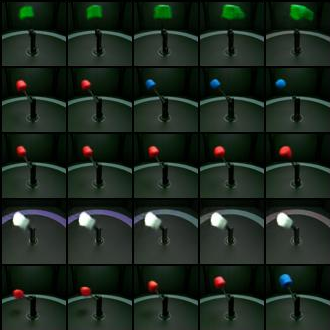} &
\includegraphics[width=0.3\textwidth]{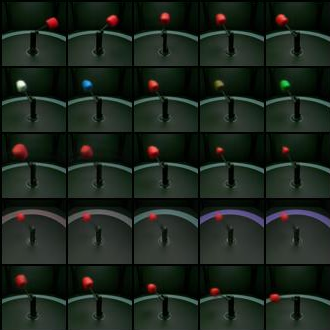} \\
\textbf{(g)}  & \textbf{(h)}  \\[6pt]
\end{tabular}
\centering
\caption{mpi3d-real dataset traverse visualization. Each row of sub-figure represents: x-rotation, object-color, size, background-color and y-rotation.
\textbf{(a)} AnnealedVAE 
\textbf{(b)} $\beta$-VAE
\textbf{(c)} FactorVAE
\textbf{(d)} DIPVAEI
\textbf{(e)} DIPVAEII
\textbf{(f)} $\beta$-TCVAE
\textbf{(g)} $\beta$-FactorTCVAE
\textbf{(h)} \textbf{SW-VAE}
}
\label{fig:mpi3d-real_traverse}
\end{figure*}

We show the traversal visualization on dSprites in \cref{fig:dsprites_traverse}, shapes3d traversal visualization in \cref{fig:shapes3d_traverse}, MPI3D-toy traversal visualization in \cref{fig:mpi3d-toy_traverse}, MPI3D-realistic traversal visualization in \cref{fig:mpi3d-realistic_traverse}, MPI3D-real traversal visualization in \cref{fig:mpi3d-real_traverse}

\end{document}